\documentclass[sigconf]{acmart}
\AtBeginDocument{%
  }

\usepackage{graphicx}
\usepackage{subcaption}

\usepackage{tabularx}
\usepackage{makecell}
\usepackage{colortbl}
\usepackage{xcolor}

\usepackage{algorithm}
\usepackage{algorithmic}

\usepackage{textcomp}

\begin{document}

\title{Mapping the Fitness Landscape: A Structure-Guided Approach to Multi-Modal Optimization}


\author{Meng Xiang}
\affiliation{%
  \institution{University of Aizu}
  \city{Aizuwakamatsu}
  \country{Japan}}
\email{mengxiang0919@gmail.com}

\author{Pei Yan}
\affiliation{%
  \institution{University of Aizu}
  \city{Aizuwakamatsu}
  \country{Japan}}
\email{peiyan@u-aizu.ac.jp}


\begin{abstract}
Multimodal optimization requires finding many optima rather than merely keeping a diverse population. Yet most niching-based evolutionary algorithms rely on distances or density estimators without explicitly recovering the underlying peak--basin organization in the decision space, which can lead to pseudo-multimodality: many distinct individuals ultimately collapse into only a few basins. We introduce Chaotic Landscape-Decoding Evolution (CLDE), a decision-space-centric framework that turns multimodal search into a closed loop of decode--value--allocate--refine. CLDE injects controlled global exploration via a logistic chaotic map with a decaying step size, then builds a $k$-nearest-neighbor graph on a decoding canvas and performs persistence-guided basin growing that merges peaks only when they are not separated by deep valleys. An adaptive persistence threshold continuously tunes the decoding resolution online to avoid over-fragmentation and over-merging. Guided by the decoded structure, CLDE carries out basin-wise selection and refinement to improve solution quality while preserving basin coverage. We instantiate CLDE as CLDE-S and CLDE-M for single- and multi-objective multimodal optimization. Experiments on 20 CEC2013 functions show that CLDE-S achieves strong peak ratio under the same evaluation budget, while on DTLZ and MMMOP suites CLDE-M attains competitive IGD/IGDx, with pronounced gains on strongly multimodal problems.
\end{abstract}



\maketitle

\section{Introduction}
\label{sec:introduction}

In fields such as robotics, drug discovery, and engineering design, multimodal optimization problems (MMOPs) aim to discover and maintain a diverse set of high-quality yet structurally distinct solutions, rather than converging to a single optimum.
Formally, a single-objective MMOP (SOMMOP) seeks a set of solutions that are well-separated in the decision space, while each solution attains sufficiently good objective quality:
\begin{equation}
\min_{i\neq j}\, \lVert x^{(i)}-x^{(j)}\rVert_{2}\ \ge\ \varepsilon_{\mathrm{dec}},
\qquad
f(x^{(i)}) \le f^\star + \varepsilon_{\mathrm{obj}},\ \ \forall i.
\label{eq:sommop_crit}
\end{equation}
On benchmarks with known optima (e.g., CEC2013), this goal is commonly instantiated by peak identification under a prescribed decision-space accuracy threshold and summarized by the peak ratio (PR).

This requirement extends to multi-objective MMOPs (MOMMOPs), where one must additionally manage Pareto trade-offs for a vector objective function $\mathbf{f}(x)$:
\begin{equation}
\min_{i\neq j}\ \lVert x^{(i)}-x^{(j)}\rVert_2\ \ge\ \varepsilon_{\mathrm{dec}},
\qquad
\max_{i\neq j}\ \big\lVert \mathbf{f}(x^{(i)})-\mathbf{f}(x^{(j)})\big\rVert_2\ \le\ \delta_{\mathrm{obj}}.
\label{eq:mommop_crit}
\end{equation}

A central difficulty in MMOPs is the information gap between the decision space and the objective space.
Since selection pressure is driven by objective values, an optimizer can be blind to decision-space topology: structurally distinct solutions may look redundant in objective space, leading to over-exploitation of a few dominant basins and the loss of important modes.
Many niching and diversity-preserving methods alleviate mode collapse, but they typically act on individuals and lack explicit mechanisms to identify stable decision-space regions and to quantify which regions deserve more search effort.
As a result, evaluations are wasted on redundant or stagnant areas, while minor but promising basins remain under-explored.

\begin{figure*}[t]
    \centering
    \includegraphics[width=0.92\linewidth, trim=50 20 30 00, clip]{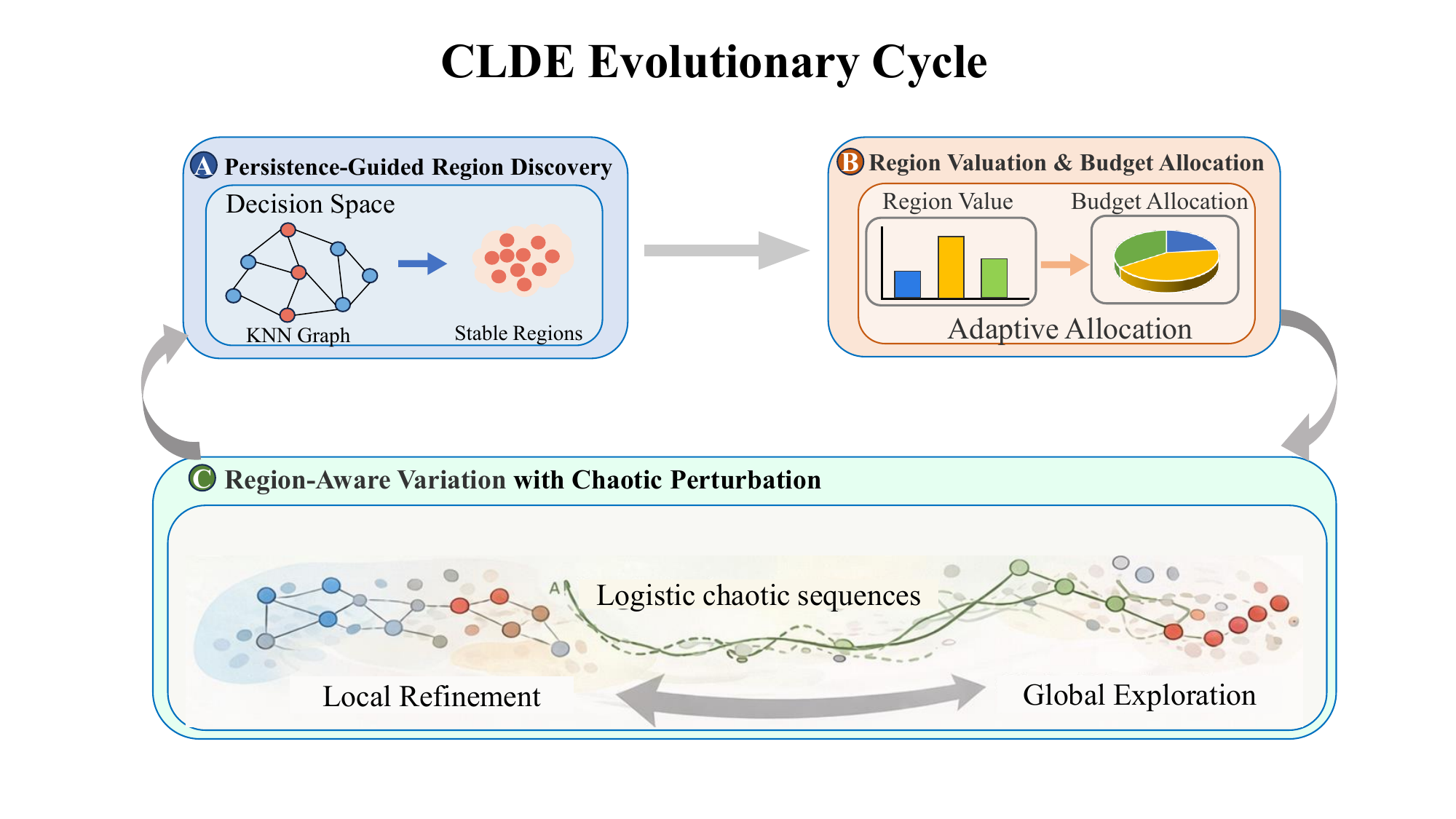}
    \caption{Overview of the CLDE evolutionary cycle with three synergistic modules.
    (A) \textbf{Persistence-guided region discovery}: build a decision-space neighborhood graph (e.g., kNN) and decode stable basins/regions by merging shallow structures.
    (B) \textbf{Region valuation \& adaptive budget allocation}: compute region values and allocate search budgets proportionally to focus evaluations on informative regions.
    (C) \textbf{Region-aware variation with chaotic perturbation}: refine solutions locally within regions while using logistic chaotic sequences to trigger controlled global exploration and cross-region jumps.}
    \label{fig:clde_cycle}
\end{figure*}

To address this root cause, we propose CLDE, a region-centric framework that turns multimodal search into a closed-loop process over decision-space regions (Figure.~\ref{fig:clde_cycle}).
CLDE iterates three steps: it decodes stable regions from sampled solutions, assigns each region a saliency score and allocates budgets accordingly, and performs region-aware variation with controlled chaotic perturbations.
This cycle bridges the structural information gap by making basin structure explicit, using region valuation for principled budget control, and enabling cross-basin transitions without sacrificing local refinement.

We instantiate CLDE for both single-objective and multi-objective multimodal optimization and evaluate it on standard benchmarks.
Results demonstrate that CLDE consistently improves mode coverage, convergence stability, and structural diversity preservation across diverse multimodal landscapes.

This paper makes the following contributions:
\begin{itemize}
  \item We propose CLDE, a region-centric framework that closes the loop over decision-space basins to bridge the decision--objective gap in multimodal optimization.
  \item We develop persistence-guided region discovery to decode stable basins from a decision-space neighborhood graph and suppress spurious/shallow modes.
  \item We introduce region saliency and adaptive budget allocation to focus evaluations on important basins while preserving minor-but-promising ones.
  \item We design a region-aware variation with logistic-chaos perturbations for controlled cross-basin exploration and within-basin refinement.
\end{itemize}

\section{Related Work}
\label{sec:related}

The proposed CLDE framework is most closely related to three research lines: fitness landscape analysis and structure decoding, single-objective multimodal evolutionary algorithms, and multimodal multi-objective evolutionary algorithms. Although these directions have developed effective mechanisms for preserving multiple modes, it remains nontrivial to (i) decode stable decision-space regions online under limited evaluations and (ii) translate such structural signals into region-level resource allocation within an evolutionary loop.

\subsection{Fitness landscape analysis and structure decoding}
Fitness landscape analysis (FLA) aims to characterize macroscopic properties of optimization problems, including ruggedness, correlation length, and the distribution of local optima, and is often used for diagnosis and algorithm comparison~\cite{zou2022survey,tan2021differential,pitzer2012fla}. Classical FLA tools (e.g., random-walk correlation and information-theoretic indicators) provide global summaries of difficulty but typically do not reconstruct basin-level structures explicitly.

To obtain a more structural view, local optima networks represent local optima as nodes and transitions between their basins as edges, revealing funnel structures and escape pathways on combinatorial landscapes~\cite{Ochoa2018}. For continuous problems, several studies estimate the number and size of basins by performing local search from random starting points and counting distinct local optima and their attraction domains~\cite{preuss2012improved,li2011multimodalSurvey}. Graph-based approaches construct neighborhood graphs over samples and analyze connectivity or contour structures to infer the existence of multiple basins~\cite{stoean2010topological}.

In parallel, topological data analysis (TDA) offers robust tools to extract features from scalar fields and point clouds. Persistent homology tracks the birth and death of connected components, loops, and voids during a filtration and separates short-lived noise from persistent structures~\cite{edelsbrunner2010computational,chazal2017tda}. For scalar-valued fitness functions, $0$-dimensional persistent homology is closely related to the persistence of connected components in excursion sets, which naturally reflects the emergence and disappearance of peaks and basins. Nevertheless, much of the above structure analysis is used in an offline manner and often assumes relatively dense sampling; integrating such structural decoding tightly into an online evolutionary process under strict evaluation budgets remains less explored.

\subsection{Single-objective multimodal evolutionary algorithms}
SOMMOP aims to locate multiple local and global optima of a scalar objective instead of a single best solution. Classical genetic niching methods include fitness sharing~\cite{goldberg1987sharing}, Mahfoud's general niching framework~\cite{mahfoud1995niching}, clearing~\cite{petrowski1996clearing}, and speciation-based genetic algorithms~\cite{li2002speciation}. These approaches introduce sharing functions or restricted competition so that crowded regions are penalized, enabling several peaks to be preserved simultaneously; comprehensive surveys summarize typical mechanisms and failure modes~\cite{li2011multimodalSurvey}.

Differential Evolution (DE) is also a popular base for multimodal search due to its simple structure and self-adaptive steps. Crowding DE applies crowding selection so that each offspring competes against its nearest neighbor, promoting niche coexistence~\cite{thomsen2004crowdingDE}. Neighborhood-based mutation and local search mechanisms have also been proposed to enhance the ability of DE to track multiple basins~\cite{qu2012deNeighborhood}. Structure- or topology-aware niching further clusters individuals by spatial distribution and maintains representatives for each species~\cite{stoean2010topological,preuss2012improved}. More recent multimodal DE variants emphasize adaptive niching and multi-scale search, including adaptive niche radii, dual-strategy designs, contour-inspired boundary inference, and multi-stage exploration--refinement schemes~\cite{chen2019novelDE,wang2018dual,wang2019automatic,wang2019multilevel}.

Chaos-driven dynamics have also been widely used to enhance exploration by introducing deterministic yet non-periodic perturbations~\cite{caponetto2003chaotic,zelinka2010evolutionary,peng2009differential,pei2014chaotic,alatas2010chaotic}. While these methods can mitigate premature convergence and mode collapse, mode control is often implemented via individual-level distance rules and local competition. Under tighter evaluation budgets, it becomes challenging to explicitly prioritize stable and informative basins over redundant or stagnant ones, motivating region-centric designs that couple structure decoding with budget control.

\subsection{Multimodal multi-objective evolutionary algorithms}
MOMMOP aims to approximate the Pareto front in the objective space while preserving multiple structurally distinct solution families in the decision space that realize similar trade-offs~\cite{deb2010multimodalMO,li2011multimodalSurvey}. Classical MOEAs such as NSGA-II~\cite{deb2002nsga2}, SPEA2~\cite{zitzler2001spea2}, and MOEA/D~\cite{zhang2007moead} mainly emphasize convergence and diversity in the objective space, and can therefore collapse to a few decision-space basins even when the objective-space front appears well covered~\cite{coello2007emoSurvey}.

To address multimodality more explicitly, a first family augments Pareto-based selection with decision-space grouping or local-front modeling. DN-NSGA-II clusters solutions in the decision space and biases mating and survival by cluster to maintain disconnected local fronts~\cite{Liu2016DNNSGAII}, while HREA detects and ranks local fronts hierarchically and injects such geometry/indicator cues into selection~\cite{Li2022HREA}. A second family relies on neighborhood- or swarm-based mechanisms to sustain multiple branches, including multi-swarm maintenance and ring-neighborhood interactions~\cite{fieldsend2014running,Yue2017MORingPSO,liang2018self}. A third family adopts decomposition, coevolution, and indicator-assisted selection to balance convergence and coverage, such as CoMMEA~\cite{Li2023CoMMEA}, RVEA~\cite{Cheng2016Reference}, MMEA-WI~\cite{Li2021MMEAWI}, and CMMO~\cite{Ming2023CMMO}.

For expensive objectives, surrogate- and classification-assisted MOMMOPs are also popular and largely orthogonal to the above mechanisms. Classification/regression models can filter candidates and reduce evaluations~\cite{zhang2015classification,pan2018classification}, while Kriging and ensemble decision models can be combined with reference-vector guidance or decomposition to improve robustness and coverage~\cite{chugh2016surrogate,lin2021ensemble,sonoda2022multiple}. Overall, MOMMOP research has substantially improved decision-space diversity, yet the maintenance of basin structure is often driven by local geometric cues or static partition rules, and explicit region-wise valuation and budget allocation tightly synchronized with the evolutionary loop remain relatively limited.

\section{Proposed Method}
\label{sec:method}

We propose \emph{Chaotic Landscape-Decoding Evolution} (CLDE), a unified framework for single-objective and multi-objective multimodal optimization.
CLDE closes an online loop over \emph{decision-space basins} rather than isolated individuals.
At each generation, (i) \textbf{chaotic evolution} produces structurally diverse candidates to avoid premature concentration, (ii) \textbf{persistence-guided decoding} reconstructs basin structure from sampled points on a neighborhood graph, and (iii) \textbf{saliency-guided allocation} distributes a limited evaluation budget across basins using structure evidence (depth and support).
The same loop is shared by both settings, while the survival rule follows the standard single-objective greedy improvement or multi-objective nondominated selection.

\subsection{Chaotic Evolution for Basin-Covering Exploration}
\label{subsec:ce}

In multimodal landscapes, small random perturbations often behave like local hill-climbing and fail to cross ridges,
whereas overly aggressive perturbations destroy late-stage refinement.
CLDE employs a bounded yet nonperiodic chaotic driver to create occasional long jumps early on and controlled refinement later,
thereby improving basin-to-basin transitions under the same evaluation budget.

We maintain a chaotic state matrix $Z^{(t)}\in(0,1)^{N\times D}$ and update it element-wise by a logistic map:
\begin{equation}
Z^{(t+1)}_{ij}=\mu Z^{(t)}_{ij}(1-Z^{(t)}_{ij}),\qquad \mu=4.
\label{eq:clde_logistic}
\end{equation}
It is mapped to a signed factor $s^{(t+1)}_{ij}=2Z^{(t+1)}_{ij}-1\in(-1,1)$, and we apply a multiplicative perturbation with a decaying step size $\eta_t$:
\begin{equation}
x'^{(t)}_{ij}=\mathrm{clip}\!\Big(x^{(t)}_{ij}\bigl(1+\eta_t s^{(t+1)}_{ij}\bigr),\,L_j,\,U_j\Big),
\qquad \eta_{t+1}=\alpha\eta_t,\ \alpha\in(0,1).
\label{eq:clde_chaotic_update}
\end{equation}

\begin{algorithm}[t]
\caption{Chaotic Evolution used in CLDE}
\label{alg:clde_ce}
\small
\begin{algorithmic}[1]
\REQUIRE Population $P^{(t)}=\{x_i^{(t)}\}_{i=1}^{N}$, bounds $[L,U]$, chaotic matrix $Z^{(t)}$, step size $\eta_t$, decay $\alpha$, crossover rate $\mathrm{Cr}$.
\ENSURE Candidate set $P'^{(t)}=\{x_i'^{(t)}\}_{i=1}^{N}$.
\STATE \textbf{Update chaotic state} $Z^{(t+1)} \leftarrow \mu Z^{(t)}\odot(1-Z^{(t)})$ by Eq.~\eqref{eq:clde_logistic}.
\FOR{$i=1$ \TO $N$}
  \STATE $x'_i \leftarrow x_i^{(t)}$; choose a pivot dimension $k\sim\mathrm{Unif}\{1,\dots,D\}$.
  \FOR{$j=1$ \TO $D$}
    \IF{$\mathrm{rand}()<\mathrm{Cr}$ \OR $j=k$}
      \STATE $s \leftarrow 2Z^{(t+1)}_{ij}-1$.
      \STATE $x'_{ij} \leftarrow \mathrm{clip}\big(x^{(t)}_{ij}(1+\eta_t s),\,L_j,\,U_j\big)$ \hfill (Eq.~\eqref{eq:clde_chaotic_update})
    \ENDIF
  \ENDFOR
\ENDFOR
\STATE $\eta_{t+1}\leftarrow \alpha\eta_t$.
\RETURN $P'^{(t)}$.
\end{algorithmic}
\end{algorithm}

\begin{figure}[t]
  \centering
  \includegraphics[width=0.95\linewidth, trim=60 60 50 60, clip]{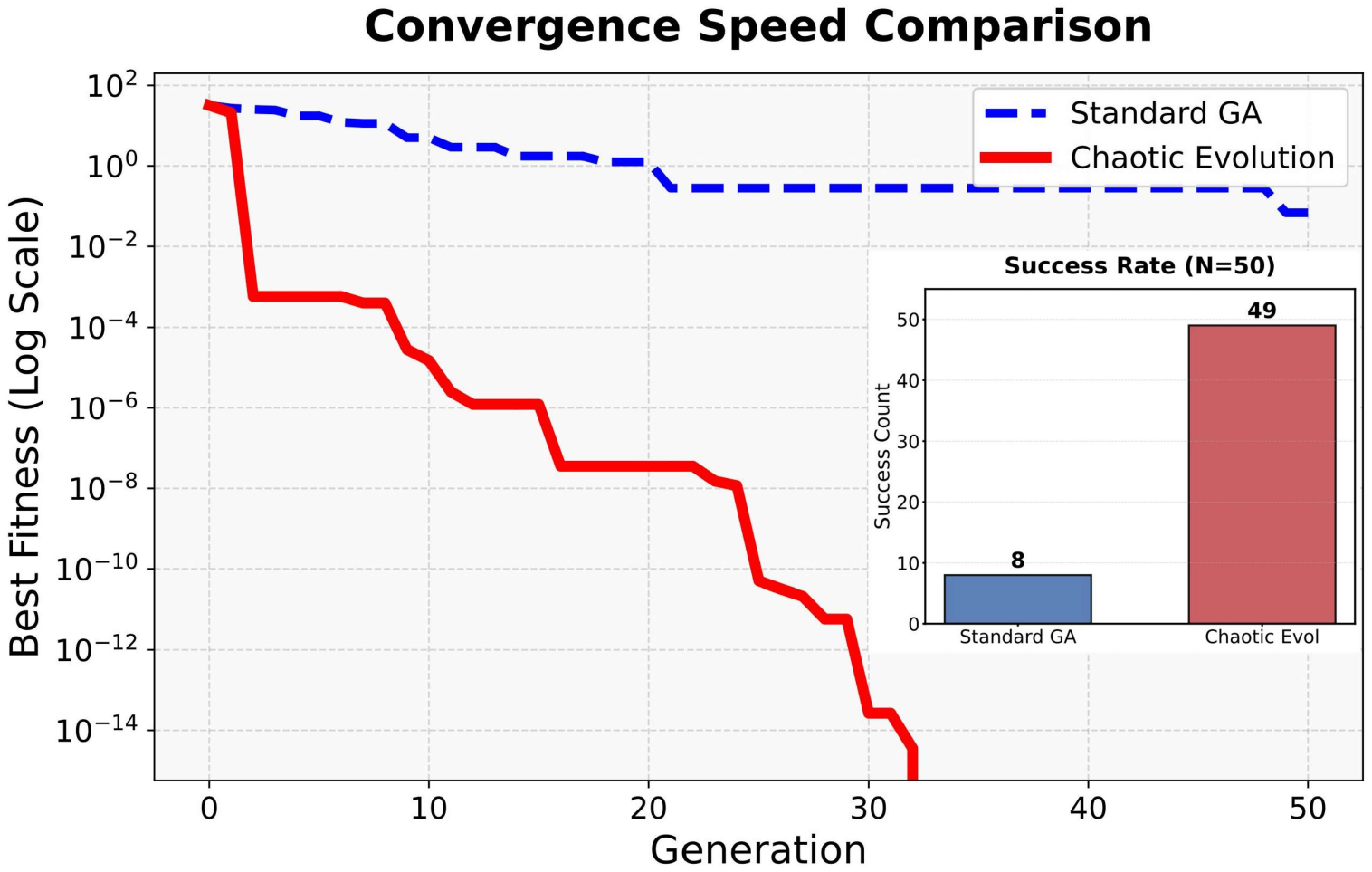}
  \caption{Convergence speed and success statistics in a worst-case escape test.
  The red curve (CE) drives the best fitness down by many orders of magnitude,
  while the blue dashed curve (standard GA) stagnates at a higher level.
  The inset shows that CE succeeds in almost all runs, whereas standard GA succeeds only rarely.}
  \label{fig:ce_convergence}
\end{figure}

Figure~\ref{fig:ce_convergence} reports the convergence and success statistics in a worst-case escape test, illustrating that CE improves ridge-crossing behavior under the same budget.

\subsection{Landscape Structure Decoding via Persistence-Guided Basin Growing}
\label{subsec:decoding}

Objective-space feedback can be blind to the decision-space basin topology, which often results in pseudo-diversity (many points in one basin) or mode deletion (losing a structurally distinct basin).
CLDE therefore decodes basin structure online from sampled points, producing explicit basin representatives and stable regions that can be budgeted and maintained in later stages.

Let $\{x_i\}_{i=1}^{n}$ denote the decoding canvas $\mathcal{S}^{(t)}=\mathcal{P}^{(t)}\cup\mathcal{P}'^{(t)}$, where $n=|\mathcal{S}^{(t)}|$.
We construct a $k$NN graph $G=(V,E)$ over the decoding canvas and define a scalar height $\tilde f(i)$ for each node.
For single-objective minimization, we set $\tilde f(i)=-f(x_i)$; for multi-objective optimization, we set $\tilde f(i)=q(x_i)$ where $q(\cdot)$ is a scalar quality derived from nondominated rank and crowding distance.
CLDE processes nodes in descending $\tilde f$ order and attaches each node to the basin of its highest processed neighbor (discrete ascent).
When multiple basins are connected at a node, a persistence-style rule merges basins if their separation is below a threshold $\tau^{(t)}$.
Algorithm~\ref{alg:clde_decode} summarizes this growing-and-merging procedure and returns a set of decoded basins $\{\mathcal{B}_k^{(t)}\}$ and their peak representatives $\{r_k^{(t)}\}$.

\begin{algorithm}[t]
\caption{Persistence-Guided Basin Decoding}
\label{alg:clde_decode}
\small
\begin{algorithmic}[1]
\REQUIRE Decoding canvas $\{x_i\}_{i=1}^{n}$, $k$NN graph $G=(V,E)$, heights $\tilde f(i)$, persistence threshold $\tau^{(t)}$.
\ENSURE Basins $\{\mathcal{B}_k^{(t)}\}$, peak representatives $\{r_k^{(t)}\}$, and basin count $K^{(t)}$.
\STATE \textbf{Sort} nodes by descending height: $\pi(1),\dots,\pi(n)$ with $\tilde f(\pi(1))\ge \cdots \ge \tilde f(\pi(n))$.
\STATE Initialize disjoint sets; mark all nodes inactive.
\FOR{$\ell=1$ \TO $n$}
  \STATE $i\leftarrow \pi(\ell)$; let $\mathcal{N}^+(i)$ be active neighbors of $i$ in $G$.
  \IF{$\mathcal{N}^+(i)=\emptyset$}
    \STATE \textbf{Create new basin} $e$ with representative $r(e)\leftarrow i$; assign $i$ to $e$.
  \ELSE
    \STATE $g(i)\leftarrow \arg\max_{j\in\mathcal{N}^+(i)} \tilde f(j)$; assign $i$ to basin of $g(i)$.
    \FORALL{basins $e$ incident to other neighbors in $\mathcal{N}^+(i)$}
      \IF{$e\neq e_{g(i)}$ \AND $\min\{\tilde f(r(e)),\tilde f(r(e_{g(i)}))\}<\tilde f(i)+\tau^{(t)}$}
        \STATE \textbf{Merge} $e$ into $e_{g(i)}$; keep higher representative:
        $r \leftarrow \arg\max\{\tilde f(r(e)),\tilde f(r(e_{g(i)}))\}$.
      \ENDIF
    \ENDFOR
  \ENDIF
  \STATE Mark $i$ active.
\ENDFOR
\STATE Extract final basins $\{\mathcal{B}_k^{(t)}\}$ and representatives $\{r_k^{(t)}\}$.
\STATE $K^{(t)}\gets$ number of decoded basins.
\RETURN $\{\mathcal{B}_k^{(t)}\}, \{r_k^{(t)}\}, K^{(t)}$.
\end{algorithmic}
\end{algorithm}

\paragraph{Adaptive persistence threshold (online resolution control).}
The threshold $\tau^{(t)}$ controls the decoding resolution: a smaller value preserves finer peak separation but can over-fragment basins under sparse/noisy samples, whereas a larger value stabilizes decoding but may over-merge distinct peaks as the search concentrates.
To keep the resolution consistent across generations, CLDE updates $\tau$ \emph{online} using the decoded basin count as a feedback signal.
Let $K^{(t)}$ be the basin count returned by Algorithm~\ref{alg:clde_decode} at generation $t$, and let $K_{\mathrm{tar}}$ be a target basin count (e.g., $K_{\mathrm{tar}}=\lfloor \sqrt{N}\rceil$, clipped to a reasonable range).
We update
\begin{equation}
\tau^{(t+1)}=\mathrm{clip}\!\left(
\tau^{(t)}\exp\!\Big(\gamma_\tau \tfrac{K^{(t)}-K_{\mathrm{tar}}}{K_{\mathrm{tar}}}\Big),
\ \tau_{\min},\tau_{\max}\right),
\label{eq:clde_tau_adapt}
\end{equation}
where $\gamma_\tau$ is the feedback gain and $\mathrm{clip}(\cdot)$ enforces $\tau^{(t)}\in[\tau_{\min},\tau_{\max}]$ for numerical stability.
Intuitively, if $K^{(t)}>K_{\mathrm{tar}}$, $\tau$ increases to suppress short-lived structures by stronger merging in the next generation; if $K^{(t)}<K_{\mathrm{tar}}$, $\tau$ decreases to retain finer basin separation.
This online resolution control makes persistence-guided decoding robust throughout the evolutionary loop and directly supports the basin-wise budgeting described next.

\paragraph{MO height by rank--crowding scalarization.}
For MOMMOP, CLDE defines the node height $\tilde f(i)$ by combining nondominated rank and crowding distance.
Let $r_i\in\{1,2,\dots\}$ be the nondominated rank of $x_i$ (smaller is better), and let $c_i\ge 0$ be its crowding distance (larger is better).
We normalize the crowding distance within the current decoding canvas as
\begin{equation}
\bar c_i=\frac{c_i-\min_j c_j}{\max_j c_j-\min_j c_j+\varepsilon}.
\label{eq:clde_crowd_norm}
\end{equation}
Then the scalar quality is defined as
\begin{equation}
q(x_i)=\frac{R_{\max}-r_i}{R_{\max}-1+\varepsilon}+\kappa\,\bar c_i,
\qquad
\tilde f(i)=q(x_i),
\label{eq:clde_rankcrowd}
\end{equation}
where $R_{\max}$ is the worst rank observed in the decoding canvas and $\kappa>0$ controls the contribution of the (normalized) crowding term.

\subsection{Saliency-Guided Basin-Wise Budget Allocation}
\label{subsec:saliency}

After decoding, a key question is how to distribute a limited per-generation budget across basins.
Uniform allocation wastes evaluations on weak/noisy basins, while aggressive pruning risks losing structurally distinct modes.
CLDE therefore assigns each decoded basin a saliency score from \emph{structural evidence} and allocates resources basin-wise, enabling both refinement of dominant basins and preservation of minor basins.

\begin{figure}[t]
  \centering
  \includegraphics[width=0.95\linewidth, trim=5 80 0 0, clip]{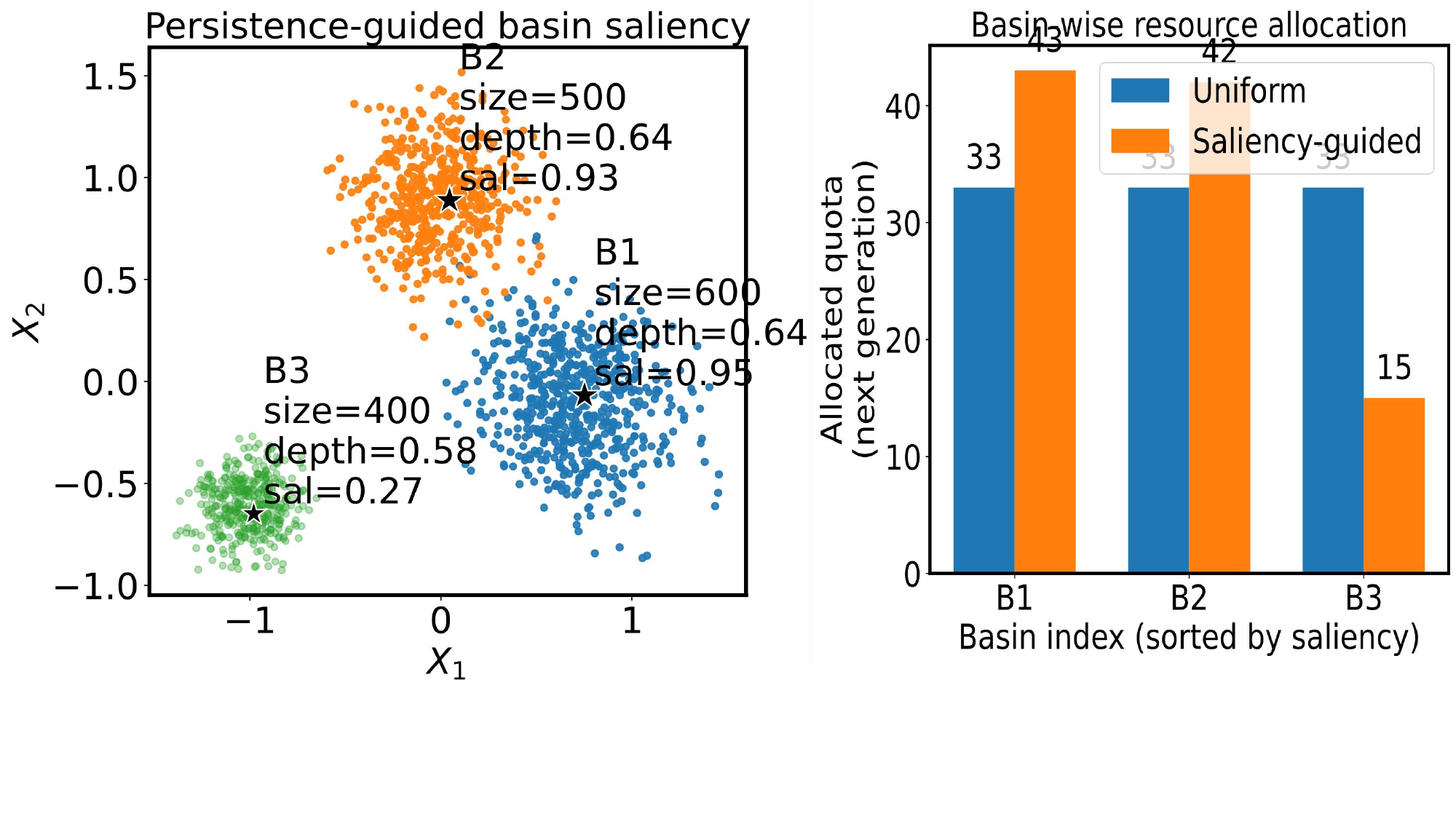}
  \caption{Persistence-guided basin saliency and resource allocation.
  Left: decoded basins in the decision space with approximate support size, persistence depth, and saliency scores.
  Right: uniform basin-wise allocation vs.\ the proposed saliency-guided quotas.}
  \label{fig:gmm_saliency}
\end{figure}

For each basin $\mathcal{B}_k^{(t)}$, we estimate (i) a depth/persistence proxy $\Delta_k$ from the height contrast within the basin and (ii) a support proxy $s_k=|\mathcal{B}_k^{(t)}|$.
After normalization to $[0,1]$, we compute
\begin{equation}
\mathrm{sal}_k \;=\; \beta\,\bar\Delta_k + (1-\beta)\,\bar s_k,\qquad \beta\in[0,1],
\label{eq:clde_saliency}
\end{equation}
and convert saliency into basin-wise quotas with a minimum quota $Q_{\min}$:
\begin{equation}
Q_k \;=\; Q_{\min} \;+\; \Big\lfloor (N-KQ_{\min})\cdot \frac{\mathrm{sal}_k}{\sum_{j=1}^{K}\mathrm{sal}_j+\varepsilon}\Big\rfloor,
\label{eq:clde_quota}
\end{equation}
where $K$ is the decoded basin count.
Figure~\ref{fig:gmm_saliency} illustrates how decoded basins are scored and how the saliency-guided allocation differs from uniform allocation.

\subsection{Unified Evolutionary Loop for SO and MO}
\label{subsec:unified}

Single-objective and multi-objective multimodal optimization share the same basin-maintenance difficulty in the decision space, but differ in survival selection.
CLDE keeps the same exploration--decoding--allocation backbone and switches only the acceptance rule, yielding a unified algorithmic template.

\begin{algorithm}[t]
\caption{CLDE: Chaotic Landscape-Decoding Evolution}
\label{alg:clde_unified}
\small
\begin{algorithmic}[1]
\REQUIRE Population size $N$, max generations $T$, bounds $[L,U]$, $k$NN parameter $k$, initial threshold $\tau^{(0)}$, min quota $Q_{\min}$.
\STATE Initialize population $\mathcal{P}^{(0)}$ of size $N$; evaluate objectives.
\FOR{$t=0$ \TO $T-1$}
  \STATE \textbf{(1) Chaotic exploration:} generate candidates $\mathcal{P}'^{(t)}$ using Algorithm~\ref{alg:clde_ce}.
  \STATE \textbf{(2) Form decoding canvas:} $\mathcal{S}^{(t)} \leftarrow \mathcal{P}^{(t)} \cup \mathcal{P}'^{(t)}$; $n \leftarrow |\mathcal{S}^{(t)}|$.
  \STATE \textbf{(3) Decode basins:} build $k$NN graph on $\mathcal{S}^{(t)}$; compute heights $\tilde f$; run Algorithm~\ref{alg:clde_decode} to obtain $\{\mathcal{B}_k^{(t)}\}$, $\{r_k^{(t)}\}$, and $K^{(t)}$.
  \STATE \textbf{(3b) Update $\tau$:} $\tau^{(t+1)} \leftarrow$ Eq.~\eqref{eq:clde_tau_adapt}.
  \STATE \textbf{(4) Saliency \& allocation:} compute $\mathrm{sal}_k$ (Eq.~\ref{eq:clde_saliency}) and real-valued quotas $Q_k^\star$ (Eq.~\ref{eq:clde_quota}).
  \STATE \textbf{(4b) Quota correction:} $Q_k \leftarrow \max(Q_{\min},\lfloor Q_k^\star \rfloor)$; $R \leftarrow N-\sum_{k=1}^{K^{(t)}} Q_k$.
  \STATE \hspace{0.8em}\textbf{if} $R>0$: add $R$ slots to basins with largest $(Q_k^\star-\lfloor Q_k^\star \rfloor)$ (ties by larger $\mathrm{sal}_k$);
  \STATE \hspace{0.8em}\textbf{else if} $R<0$: remove $|R|$ slots from basins with smallest $(Q_k^\star-\lfloor Q_k^\star \rfloor)$ (ties by smaller $\mathrm{sal}_k$);
  \STATE \hspace{0.8em}ensure $Q_k\ge Q_{\min}$ and $\sum_{k=1}^{K^{(t)}} Q_k=N$.
  \STATE \textbf{(5) Basin-wise update:}
  \FOR{$k=1$ \TO $K^{(t)}$}
    \STATE Select $Q_k$ candidates from $\mathcal{B}_k^{(t)}$ for survival/variation.
    \STATE \textcolor{red}{\textbf{SO:} greedy elitism within each basin.}
    \STATE \textcolor{blue}{\textbf{MO:} buffer candidates for global nondominated selection.}
  \ENDFOR
  \STATE \textcolor{red}{\textbf{SO:} $\mathcal{P}^{(t+1)} \leftarrow$ union of basin elites (sizes $\{Q_k\}$).}
  \STATE \textcolor{blue}{\textbf{MO:} $\mathcal{P}^{(t+1)} \leftarrow$ Select$_N$ by nondominated sorting + crowding distance.}
\ENDFOR
\STATE \textbf{Return} \textcolor{red}{multimodal solution set (SO)} \textbf{or} \textcolor{blue}{multi-modal Pareto set (MO)}.
\end{algorithmic}
\end{algorithm}

\section{Experimental}
\label{sec:exp}

In this section, we empirically evaluate the proposed chaotic landscape-decoding framework in both its single-objective instantiation (denoted \textbf{CLDE-S}) and multi-objective instantiation (denoted \textbf{CLDE-M}). 

\subsection{Experimental Setup}\label{sec:exp_setup_full}
Table~\ref{tab:clde_params} summarizes the key hyperparameters of CLDE framework together with their default values and the corresponding components in Sec.~\ref{sec:method}. 
All experiments in this paper use the default settings in Table~\ref{tab:clde_params} to ensure consistent and fair comparisons across benchmarks, and all reported results are averaged over 30 independent runs.

\begin{table}[t]
\centering
\caption{Key parameters in the Chaotic Landscape--Decoding Evolution (CLDE) framework.}
\label{tab:clde_params}
\footnotesize
\setlength{\tabcolsep}{3pt}
\renewcommand{\arraystretch}{1.05}
\begin{tabularx}{\columnwidth}{@{}l c X@{}}
\toprule
\textbf{Parameter (code name)} & \textbf{Default} & \textbf{Role (used in Sec.~\ref{sec:method})} \\
\midrule
\texttt{max\_generations} ($T$)      & 200  & Generations (Alg.~\ref{alg:clde_unified}). \\
\texttt{population\_size} ($N$)      & 100  & Population / total budget (Eq.~\ref{eq:clde_quota}). \\

\texttt{chaotic\_mu} ($\mu$)         & 4.0  & Logistic map (Eq.~\ref{eq:clde_logistic}). \\
\texttt{chaotic\_step\_init} ($\eta_0$) & 0.5  & Init step (Eq.~\ref{eq:clde_chaotic_update}). \\
\texttt{chaotic\_step\_decay} ($\alpha$)   & 0.99 & Step decay (Eq.~\ref{eq:clde_chaotic_update}). \\
\texttt{crossover\_rate} ($\mathrm{Cr}$)   & 0.9  & Update prob.\ (Alg.~\ref{alg:clde_ce}). \\

\texttt{canvas\_size} ($n$)            & 100  & Decoding canvas (Sec.~\ref{subsec:decoding}). \\
\texttt{k\_neighbors} ($k$)          & 10   & Graph neighbors (Alg.~\ref{alg:clde_decode}). \\

\texttt{persistence\_tau\_init} ($\tau^{(0)}$) & 0.10 & Init merge level (Alg.~\ref{alg:clde_decode}). \\
\texttt{tau\_bounds\_gain} ($\tau_{\min},\tau_{\max},\gamma_\tau$) & (0.02, 0.30, 0.20) & $\tau$ range / gain (Eq.~\ref{eq:clde_tau_adapt}). \\

\texttt{saliency\_beta} ($\beta$)     & 0.70 & Depth--size mix (Eq.~\ref{eq:clde_saliency}). \\
\texttt{quota\_min} ($Q_{\min}$)      & 1    & Min quota (Eq.~\ref{eq:clde_quota}). \\

\texttt{rankcrowd\_kappa} ($\kappa$)  & 1.0  & Rank--crowd scale (Eq.~\ref{eq:clde_rankcrowd}). \\
\texttt{local\_sigma} ($\sigma_{\text{loc}}$) & 0.05 & Local refine (Alg.~\ref{alg:clde_unified}). \\
\texttt{archive\_size}               & 5    & Archive cap.\ for basin-wise maintenance (Sec.~\ref{subsec:unified}). \\
\bottomrule
\end{tabularx}
\end{table}

\subsection{Single-Objective Multimodal Optimization}
\subsubsection{Benchmark functions and evaluation metric}
For single-objective multimodal optimization, we adopt the 20 CEC2013 test functions (F1--F20).
For the complete benchmark list and detailed specifications, please refer to the Supplementary Material (Sec.~1.1, Table~S1).

Following the CEC2013 guidelines, we use the Peak Ratio (PR) as the primary performance indicator:
\begin{equation}
  \mathrm{PR}
  = \frac{\sum_{run = 1}^{NR} NPF_{run}}{NKP \cdot NR},
  \label{con:PR}
\end{equation}
where $NPF_{run}$ is the number of global optima located in a run, $NKP$ is the known number of global optima, and $NR$ is the number of independent runs. We adopt the CEC2013 accuracy threshold of $1.0\times10^{-4}$ for counting a global optimum as found.

\begin{table}[t]
\centering
\caption{Peak ratio comparison of CLDE-S and state-of-the-art algorithms on CEC2013 benchmarks (F1--F20) under the decision-space accuracy threshold $1.0\times10^{-4}$ for peak identification.}
\label{table:PR}
\setlength{\tabcolsep}{4pt}
\footnotesize
\begin{tabular}{@{}lcccccc@{}}
\toprule
No. & CLDE-S & ANDE & MM-DE & NEA2 & NMMSO & dADE \\
\midrule
F1  & \textbf{1.00} & \textbf{1.00} & \textbf{1.00} & \textbf{1.00} & \textbf{1.00} & \textbf{1.00} \\
F2  & \textbf{1.00} & \textbf{1.00} & \textbf{1.00} & \textbf{1.00} & \textbf{1.00} & \textbf{1.00} \\
F3  & \textbf{1.00} & \textbf{1.00} & \textbf{1.00} & \textbf{1.00} & \textbf{1.00} & \textbf{1.00} \\
F4  & \textbf{1.00} & \textbf{1.00} & \textbf{1.00} & \textbf{1.00} & \textbf{1.00} & \textbf{1.00} \\
F5  & \textbf{1.00} & \textbf{1.00} & \textbf{1.00} & \textbf{1.00} & \textbf{1.00} & \textbf{1.00} \\
F6  & \textbf{1.00} & \textbf{1.00} & \textbf{1.00} & 0.960 & 0.660 & 0.990 \\
F7  & \textbf{1.000} & 0.937 & 0.916 & 0.945 & \textbf{1.000} & 0.880 \\
F8  & \textbf{0.986} & 0.946 & 0.971 & 0.240 & 0.897 & 0.989 \\
F9  & 0.968 & 0.511 & 0.463 & 0.621 & 0.978 & 0.828 \\
F10 & \textbf{1.000} & \textbf{1.000} & \textbf{1.000} & \textbf{1.000} & \textbf{1.000} & \textbf{1.000} \\
F11 & 0.982 & \textbf{1.000} & \textbf{1.000} & 0.980 & 0.990 & 0.866 \\
F12 & 0.861 & \textbf{1.000} & \textbf{1.000} & 0.852 & 0.993 & 0.748 \\
F13 & 0.854 & 0.714 & 0.667 & 0.976 & 0.983 & 0.736 \\
F14 & 0.667 & 0.666 & 0.666 & \textbf{0.830} & 0.721 & 0.667 \\
F15 & \textbf{0.750} & 0.636 & \textbf{0.750} & 0.742 & 0.635 & 0.642 \\
F16 & 0.620 & 0.667 & 0.667 & 0.673 & 0.660 & \textbf{0.890} \\
F17 & \textbf{0.748} & 0.397 & 0.636 & 0.695 & 0.466 & 0.662 \\
F18 & \textbf{0.667} & 0.653 & 0.658 & \textbf{0.666} & 0.650 & 0.663 \\
F19 & 0.660 & 0.363 & 0.500 & \textbf{0.667} & 0.448 & 0.495 \\
F20 & \textbf{0.487} & 0.249 & 0.088 & 0.362 & 0.172 & 0.080 \\
\midrule
AVG & \textbf{0.862} & 0.787 & 0.799 & 0.810 & 0.813 & 0.807 \\
\bottomrule
\end{tabular}
\end{table}

\subsubsection{Comparison with state-of-the-art algorithms}
We compare the proposed CLDE-S with representative multimodal evolutionary algorithms, including NEA2~\cite{preuss2012improved}, NMMSO~\cite{fieldsend2014running}, ANDE~\cite{wang2019automatic}, MM-DE~\cite{wang2019multilevel}, and dADE.
Table~\ref{table:PR} reports the PR results on F1--F20.
CLDE-S achieves the highest overall average PR (0.862) and obtains the best or tied-best PR on the majority of functions.
On low-dimensional and structurally clear landscapes (e.g., F1--F6 and F10), CLDE-S consistently reaches $\mathrm{PR}=1.0$, indicating reliable recovery and maintenance of all global optima across runs.
On more demanding peak-rich and composite functions, CLDE-S remains among the top performers, suggesting improved basin coverage under the same evaluation budget.

\begin{table}[t]
\centering
\caption{Wilcoxon signed-rank test results between CLDE-S and other algorithms on PR over F1--F20. All comparisons are statistically significant ($p < 0.05$).}
\label{tab:wilcoxon_SO1_all_checkmark}
\setlength{\tabcolsep}{6pt}
\renewcommand{\arraystretch}{1.1}
\footnotesize
\begin{tabular}{lcc lcc}
\toprule
\textbf{Algorithm} & \textbf{$p$-value} & \textbf{Sig.} &
\textbf{Algorithm} & \textbf{$p$-value} & \textbf{Sig.} \\
\midrule
CLDE-S & -- & --  & ANDE        & 0.0196 & Yes\\
MM-DE       & 0.0422 & Yes & NEA2        & 0.0481 & Yes \\
NMMSO       & 0.0467 & Yes & dADE        & 0.0116 & Yes \\
\bottomrule
\end{tabular}
\end{table}

To verify that the improvements are not due to random variation, we further conduct Wilcoxon signed-rank tests on PR across F1--F20.
As summarized in Table~\ref{tab:wilcoxon_SO1_all_checkmark}, CLDE-S shows statistically significant advantages over all competitors ($p<0.05$), supporting the robustness of the gains in Table~\ref{table:PR}.

\subsubsection{Ablation study}
We conduct an ablation study to evaluate the contribution of each module in CLDE-S under the same protocol.
We compare: (i) CLDE-S (full), (ii) w/o Chaos (replace chaotic update with isotropic Gaussian step), (iii) w/o Persist (replace persistence-guided decoding with distance-based niching), and (iv) w/o Struct (remove the structural term from the selection score).
Table~\ref{tab:pr_ablation} reports PR results.

\begin{table}[t]
\centering
\caption{Peak ratio of CLDE-S and its ablation variants on CEC2013 benchmarks.
Here, ``w/o'' denotes CLDE-S \emph{without} the corresponding component.}
\label{tab:pr_ablation}
\setlength{\tabcolsep}{5pt}
\footnotesize
\begin{tabular}{lcccc}
\toprule
\textbf{Benchmark} & \textbf{CLDE-S} & \textbf{w/o Chaos} & \textbf{w/o Persist} & \textbf{w/o Struct} \\
\midrule
F1  & \textbf{1.000} & \textbf{1.000} & \textbf{1.000} & 0.990 \\
F2  & \textbf{1.000} & \textbf{1.000} & \textbf{1.000} & 0.990 \\
F3  & \textbf{1.000} & \textbf{1.000} & \textbf{1.000} & 0.985 \\
F4  & \textbf{1.000} & \textbf{1.000} & \textbf{1.000} & 0.985 \\
F5  & \textbf{1.000} & \textbf{1.000} & 0.995 & 0.980 \\
F6  & \textbf{1.000} & 0.990 & 0.975 & 0.940 \\
F7  & \textbf{1.000} & 0.985 & 0.965 & 0.920 \\
F8  & \textbf{0.986} & 0.950 & 0.910 & 0.780 \\
F9  & \textbf{0.968} & 0.935 & 0.895 & 0.760 \\
F10 & \textbf{1.000} & \textbf{1.000} & 0.995 & 0.970 \\
F11 & \textbf{0.982} & 0.945 & 0.900 & 0.750 \\
F12 & \textbf{0.861} & 0.820 & 0.770 & 0.600 \\
F13 & \textbf{0.854} & 0.810 & 0.760 & 0.580 \\
F14 & \textbf{0.667} & 0.630 & 0.580 & 0.420 \\
F15 & \textbf{0.750} & 0.710 & 0.660 & 0.470 \\
F16 & \textbf{0.620} & 0.590 & 0.540 & 0.380 \\
F17 & \textbf{0.748} & 0.700 & 0.645 & 0.450 \\
F18 & \textbf{0.667} & 0.620 & 0.570 & 0.390 \\
F19 & \textbf{0.660} & 0.610 & 0.560 & 0.380 \\
F20 & \textbf{0.487} & 0.430 & 0.380 & 0.250 \\
\midrule
\textbf{AVG} & \textbf{0.8625} & 0.8363 & 0.8050 & 0.6985 \\
\bottomrule
\end{tabular}
\end{table}

Overall, the full CLDE-S achieves the best performance (AVG PR $=0.8625$), followed by w/o Chaos ($0.8363$), w/o Persist ($0.8050$), and w/o Struct ($0.6985$).
The modest drop of w/o Chaos suggests that chaotic exploration mainly improves basin reachability (especially on peak-rich or higher-dimensional functions), while persistence-guided decoding provides more reliable basin separation than distance-only niching.
Removing the structural term (w/o Struct) causes the largest degradation, indicating that structure-aware selection is critical to prevent population collapse into a few dominant basins and to maintain genuine decision-space diversity.

\begin{figure*}[t]
  \centering
  \includegraphics[trim=0 270 0 0, clip, width=0.95\linewidth]{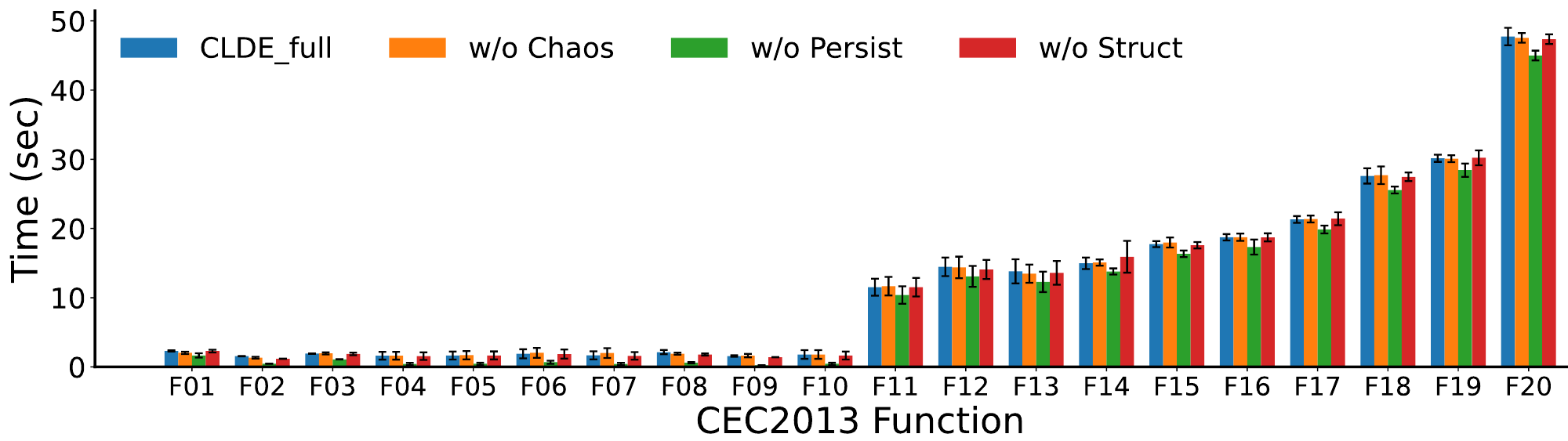}
  \caption{Runtime comparison on CEC2013 (F1--F20).
  We report the wall-clock time of CLDE\_full and three ablation variants (w/o Chaos, w/o Persist, w/o Struct); error bars denote variability across repeated runs.
  Runtime is dominated by problem complexity, while the additional decoding and structure-aware mechanisms introduce only modest overhead.}
  \label{fig:runtimeplot}
\end{figure*}

\subsubsection{Runtime analysis}
Figure~\ref{fig:runtimeplot} compares the wall-clock runtime of CLDE\_full and ablation variants under the default settings in Table~\ref{tab:clde_params}.
Runtime is largely driven by problem complexity, while the differences among configurations are relatively small.
Among the variants, w/o Persist is slightly faster, indicating that persistence-guided decoding contributes most of the additional overhead, whereas removing chaotic exploration or the structural term has only marginal impact on runtime under the same evaluation budget.

\paragraph{Decision-space visualizations.}
To avoid interrupting the main quantitative comparisons, the decision-space population snapshots that illustrate basin coverage on representative 2-D CEC2013 functions are provided in the supplementary material (Sec.~1.2, Figure.~S1)

\subsection{Multi-Objective Multimodal Optimization}
\label{subsec:mo_exp}

\subsubsection{Benchmark functions and evaluation metrics}
In the multi-objective setting, we evaluate CLDE-M on two complementary benchmark suites.
(i) \textbf{DTLZ1--DTLZ6} assess scalability on many-objective problems with diverse front geometries.
(ii) \textbf{MMMOP1--MMMOP6} explicitly feature multiple Pareto sets, which is critical for multi-objective multimodal evaluation.
The complete benchmark specifications are provided in the Supplementary Material (Sec.~2, Table~S2).
All algorithms are run under the same evaluation budget with standard settings in the literature.

As the primary objective-space indicator, we use the Inverted Generational Distance (IGD):
\begin{equation}
  \label{eq:igd}
  \mathrm{IGD}(\mathcal{A}, \mathcal{P}^\ast)
  = \frac{1}{|\mathcal{P}^\ast|}
    \sum_{z^\ast \in \mathcal{P}^\ast}
    \min_{z \in \mathcal{A}} \bigl\| z - z^\ast \bigr\|_2 .
\end{equation}
Lower IGD indicates that the approximation set is closer to and better distributed along the true Pareto front.

To assess decision-space multimodality, we additionally report IGD$_x$ on MMMOP:
\begin{equation}
  \label{eq:igdx}
  \mathrm{IGD}_x(\mathcal{A}_x, \mathcal{S}^\ast)
  = \frac{1}{|\mathcal{S}^\ast|}
    \sum_{x^\ast \in \mathcal{S}^\ast}
    \min_{x \in \mathcal{A}_x} \bigl\| x - x^\ast \bigr\|_2 .
\end{equation}
Lower IGD$_x$ means better coverage of the true Pareto set in the decision space, and thus stronger preservation of multiple Pareto-optimal families.

\subsubsection{Performance on DTLZ benchmark problems}
We compare CLDE-M with several state-of-the-art algorithms on DTLZ1--DTLZ6 using IGD; results are summarized in Table~\ref{tab:IGDmetrics}.
Overall, CLDE-M is most competitive on DTLZ1, DTLZ3, and DTLZ6, where multimodality and local fronts are pronounced, indicating that basin decoding helps avoid collapsing to a few local Pareto branches.
On smoother cases such as DTLZ2 and DTLZ4, CLDE-M remains comparable to strong baselines, suggesting that the structural mechanisms do not harm convergence when decision-space structure is simpler.

\begin{table}[t]
\centering
\caption{IGD results on DTLZ1--DTLZ6 (lower is better).}
\label{tab:IGDmetrics}
\scriptsize
\setlength{\tabcolsep}{2.0pt}
\renewcommand{\arraystretch}{1.05}
\resizebox{\columnwidth}{!}{%
\begin{tabular}{@{}lcccccc@{}}
\toprule
\textbf{Problem} & CPS~\cite{zhang2015classification} & K-RV~\cite{chugh2016surrogate} & CSEA~\cite{pan2018classification} & EDN~\cite{lin2021ensemble} & MCEA~\cite{sonoda2022multiple} & CLDE-M \\
\midrule
DTLZ1 (30) & $6.1231{\times}10^{1}$ & $5.9574{\times}10^{1}$ & $5.3458{\times}10^{1}$ & $7.6109{\times}10^{1}$ & $3.7268{\times}10^{1}$ & $\mathbf{3.4836{\times}10^{1}}$ \\
DTLZ1 (50) & $1.1710{\times}10^{2}$ & $1.2859{\times}10^{2}$ & $1.1088{\times}10^{2}$ & $1.4066{\times}10^{2}$ & $\mathbf{4.0245{\times}10^{1}}$ & $5.3412{\times}10^{1}$ \\
\midrule
DTLZ2 (30) & $1.2068$ & $1.2689$ & $8.4305{\times}10^{-1}$ & $1.3224$ & $4.3816{\times}10^{-1}$ & $\mathbf{4.2104{\times}10^{-1}}$ \\
DTLZ2 (50) & $1.8985$ & $2.9998$ & $1.7832$ & $2.8300$ & $6.2911{\times}10^{-1}$ & $\mathbf{5.7669{\times}10^{-1}}$ \\
\midrule
DTLZ3 (30) & $1.6564{\times}10^{3}$ & $1.6672{\times}10^{3}$ & $1.3159{\times}10^{3}$ & $2.1064{\times}10^{3}$ & $7.2601{\times}10^{2}$ & $\mathbf{1.7997{\times}10^{2}}$ \\
DTLZ3 (50) & $3.1099{\times}10^{3}$ & $3.5867{\times}10^{3}$ & $2.7884{\times}10^{3}$ & $3.8988{\times}10^{3}$ & $\mathbf{1.5979{\times}10^{3}}$ & $3.4269{\times}10^{2}$ \\
\midrule
DTLZ4 (30) & $1.5812$ & $1.6870$ & $9.3992{\times}10^{-1}$ & $1.4471$ & $\mathbf{8.0266{\times}10^{-1}}$ & $1.2568$ \\
DTLZ4 (50) & $2.2624$ & $3.2670$ & $2.0010$ & $2.9380$ & $\mathbf{9.0809{\times}10^{-1}}$ & $1.2946$ \\
\midrule
DTLZ5 (30) & $1.2256$ & $1.2618$ & $8.5359{\times}10^{-1}$ & $1.3895$ & $\mathbf{4.0083{\times}10^{-1}}$ & $4.8351{\times}10^{-1}$ \\
DTLZ5 (50) & $1.9899$ & $2.9974$ & $1.7169$ & $2.8174$ & $\mathbf{6.3098{\times}10^{-1}}$ & $7.7643{\times}10^{-1}$ \\
\midrule
DTLZ6 (30) & $1.9018{\times}10^{1}$ & $2.1726{\times}10^{1}$ & $2.0974{\times}10^{1}$ & $2.2788{\times}10^{1}$ & $1.2599{\times}10^{1}$ & $\mathbf{1.0894{\times}10^{1}}$ \\
DTLZ6 (50) & $3.4604{\times}10^{1}$ & $3.9568{\times}10^{1}$ & $3.8241{\times}10^{1}$ & $4.0564{\times}10^{1}$ & $2.4848{\times}10^{1}$ & $\mathbf{2.4152{\times}10^{1}}$ \\
\bottomrule
\end{tabular}%
}
\end{table}

\paragraph{MOMMOP benchmarks with decision-space evaluation}
DTLZ mainly reflects objective-space convergence/diversity and does not explicitly test whether multiple Pareto sets are preserved in the decision space.
We therefore further evaluate CLDE-M on MMMOP1--MMMOP6 and report both IGD and IGD$_x$ in Tables~\ref{tab:IGD_MMMOP} and~\ref{tab:IGDX_MMMOP}.
Together, the two metrics distinguish ``objective-space good but decision-space collapsed'' from ``good in both spaces'' outcomes.

\begin{table}[t]
\centering
\caption{IGD results on MMMOP1--MMMOP6 (lower is better).}
\label{tab:IGD_MMMOP}
\tiny
\setlength{\tabcolsep}{2.2pt}
\renewcommand{\arraystretch}{1.05}
\resizebox{\columnwidth}{!}{%
\begin{tabular}{lcccccc}
\toprule
Func. & CMMO~\cite{Ming2023CMMO} & CoMMEA~\cite{Li2023CoMMEA} & HREA~\cite{Li2022HREA} & MMEAWI~\cite{Li2021MMEAWI} & MMEAPSL~\cite{ming2024growing} & CLDE-M \\
\midrule
MMMOP1 & 5.8781e-3 & 6.5107e-3 & 9.2322e-3 & 1.1390e-2 & 6.2674e-3 & \textbf{3.9370e-3} \\
MMMOP2 & 6.4936e-3 & 2.3119e-2 & 1.7443e-2 & 8.0618e-2 & 6.9625e-3 & \textbf{4.1315e-3} \\
MMMOP3 & 6.9662e-3 & 1.8505e-2 & 1.3776e-2 & 6.9641e-3 & 6.7554e-3 & \textbf{4.2075e-3} \\
MMMOP4 & 7.1138e-3 & 5.8246e-3 & 9.7849e-3 & 1.1289e-2 & 6.9562e-3 & \textbf{4.5165e-3} \\
MMMOP5 & 6.6820e-3 & 6.4716e-3 & 8.9584e-3 & 9.4099e-3 & 6.7855e-3 & \textbf{4.3118e-3} \\
MMMOP6 & 1.9510e-2 & 1.9630e-2 & 6.2520e-2 & 3.0279e-2 & 3.8503e-2 & 1.9616e-2 \\
\bottomrule
\end{tabular}%
}
\end{table}

\begin{table}[t]
\centering
\caption{IGD$_x$ results on MMMOP1--MMMOP6 (lower is better). Baselines are the same as Table~\ref{tab:IGD_MMMOP}.}
\label{tab:IGDX_MMMOP}
\tiny
\setlength{\tabcolsep}{2.2pt}
\renewcommand{\arraystretch}{1.05}
\resizebox{\columnwidth}{!}{%
\begin{tabular}{lcccccc}
\toprule
Func. & CMMO~\cite{Ming2023CMMO} & CoMMEA~\cite{Li2023CoMMEA} & HREA~\cite{Li2022HREA} & MMEAWI~\cite{Li2021MMEAWI} & MMEAPSL~\cite{ming2024growing} & CLDE-M \\
\midrule
MMMOP1 & 6.8344e-2 & 5.7483e-2 & 6.7235e-2 & 1.5683e-1 & 8.5278e-2 & \textbf{5.7275e-2} \\
MMMOP2 & 6.0449e-2 & 2.1924e-2 & \textbf{1.3041e-2} & 1.2422e-1 & 3.2485e-2 & 3.3085e-2 \\
MMMOP3 & 1.6884e-2 & 4.4097e-2 & 2.8955e-2 & 1.9841e-2 & 1.7839e-2 & \textbf{9.2210e-3} \\
MMMOP4 & 2.9179e-2 & 5.5873e-2 & \textbf{2.2235e-2} & 3.5234e-2 & 1.7029e-2 & 2.3678e-2 \\
MMMOP5 & 8.7969e-3 & \textbf{5.7276e-3} & 1.9884e-2 & 4.1020e-2 & 8.8502e-3 & \textbf{8.0723e-3} \\
MMMOP6 & 2.8085e-1 & 8.3062e-2 & 2.3796e-1 & 3.1557e-1 & 2.8818e-1 & \textbf{7.9980e-2} \\
\bottomrule
\end{tabular}%
}
\end{table}

From Table~\ref{tab:IGD_MMMOP}, CLDE-M achieves the lowest IGD on MMMOP1--MMMOP5 and remains competitive on MMMOP6, indicating strong objective-space approximation.
More importantly, Table~\ref{tab:IGDX_MMMOP} shows that CLDE-M also attains the lowest IGD$_x$ on several problems (e.g., MMMOP1/3/6), suggesting better coverage of multiple Pareto-optimal families in the decision space.
In contrast, some baselines can reach reasonable IGD while exhibiting much worse IGD$_x$, indicating decision-space collapse that is not revealed by objective-space metrics alone.

\begin{figure}[t]
    \centering
    \includegraphics[trim=10 170 10 10, clip, width=0.95\linewidth]{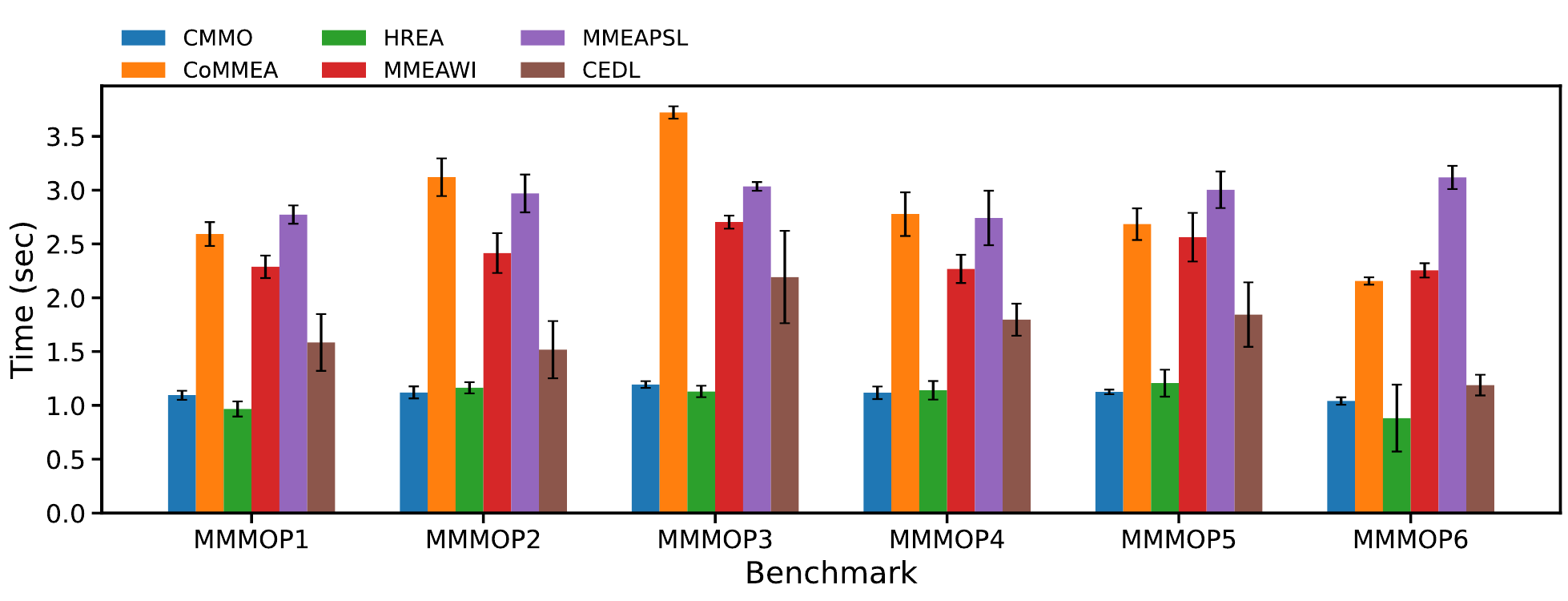}
    \caption{Average runtime (mean $\pm$ std) on MMMOP1--MMMOP6 under the same evaluation budget.
    The comparable runtime indicates that the decision-space gains of CLDE-M are not achieved by extra computational overhead.}
    \label{fig:MMMOP_runtime}
\end{figure}

Figure~\ref{fig:MMMOP_runtime} further compares runtime on MMMOP1--MMMOP6.
Since all methods share the same evaluation budget, runtimes are generally comparable, suggesting that the improved IGD$_x$ comes from the proposed decision-space mechanism rather than extra wall-clock cost.

The results are consistent with Table~\ref{tab:IGDX_MMMOP}. CLDE-M shows stable advantages on multiple MMMOP problems, supporting that the decision-space improvements are not incidental.

Taken together, the DTLZ and MMMOP results demonstrate that CLDE-M can achieve competitive objective-space convergence (IGD) while better preserving multiple decision-space Pareto families (IGD$_x$), validating the benefit of explicit basin decoding and basin-wise evolution for multimodal multi-objective optimization.

\subsection{Discussions}
\label{chap:RAA}

Overall, the results support the effectiveness of the proposed landscape--decoding framework for both single-objective and multi-objective multimodal optimization under limited evaluation budgets.

On CEC2013 single-objective benchmarks, CLDE-S attains the best overall peak ratio and achieves perfect PR on most low- and medium-dimensional functions. This suggests that combining chaotic exploration with an explicit basin view (persistence-guided decoding and structure-aware refinement) helps discover and maintain genuinely distinct optima. The Wilcoxon signed-rank tests in Table~\ref{tab:wilcoxon_SO1_all_checkmark} further confirm that the improvements over representative baselines are statistically significant.

In the multi-objective setting, DTLZ1--DTLZ6 mainly reflect objective-space convergence and diversity, and IGD is therefore an appropriate primary indicator. CLDE-M is particularly competitive on rugged or strongly multimodal cases (e.g., DTLZ1/DTLZ3/DTLZ6), indicating that decoding decision-space basins can mitigate premature convergence to a few local fronts and help maintain multiple Pareto branches. On smoother cases (e.g., DTLZ2/DTLZ4/DTLZ5), CLDE-M remains comparable to strong competitors, suggesting that the basin mechanism does not systematically harm objective-space convergence.

To evaluate decision-space multimodality more directly, we further test MMMOP1--MMMOP6 and report both IGD and IGD$_x$. Table~\ref{tab:IGD_MMMOP} shows that CLDE-M achieves the best IGD on five out of six problems (MMMOP1--MMMOP5) and remains competitive on MMMOP6. More importantly, Table~\ref{tab:IGDX_MMMOP} shows that CLDE-M achieves the lowest IGD$_x$ on MMMOP1, MMMOP3, and MMMOP6 and is competitive elsewhere, indicating better preservation of multiple decision-space solution families in several representative cases. In contrast, some baselines can obtain reasonable IGD while exhibiting much worse IGD$_x$, implying decision-space collapse that is not revealed by objective-space metrics alone.

Figure~\ref{fig:MMMOP_runtime} reports runtime on MMMOP1--MMMOP6 under the same evaluation budget. The comparable runtimes suggest that the decision-space gains are not achieved by extra computational overhead but by basin-aware search and allocation.

\section{Conclusion}
\label{chap:conclusion}
This paper presents CLDE, a decision-space-centric evolutionary framework for multimodal optimization. CLDE decodes an online peak--basin decomposition from the evolving sample set and exploits this basin view to drive selection and basin-wise resource allocation. It couples a chaotic sampling mechanism with a decaying step-size schedule to encourage early basin crossing and later-stage stabilization, applies persistence-guided basin growing on a neighborhood graph to recover a compact and stable basin structure, and performs basin-wise refinement to balance solution quality with structural coverage. A central design choice is to treat the persistence threshold $\tau$ as a resolution control and adapt it online, which mitigates both over-fragmentation and over-merging and improves decoding stability under tight evaluation budgets.

CLDE-S achieves strong peak coverage on the CEC2013 single-objective multimodal benchmarks compared with representative multimodal evolutionary algorithms. In the multi-objective setting, CLDE-M is competitive on DTLZ in terms of IGD, and results on MMMOP further suggest improved preservation of multiple decision-space Pareto families as reflected by IGD$_x$, while maintaining accurate objective-space approximation.

\bibliographystyle{ACM-Reference-Format}
\bibliography{sample-base}

\clearpage
\section*{Supplementary materials}

\section{Single-Objective Multimodal Optimization}
\label{sec:supp_so}

\subsection{Benchmark Definitions}
\label{sec:supp_benchmarks}

\noindent\textbf{Single-objective multimodal benchmarks.}
We use 20 widely adopted single-objective multimodal test functions (Table~\ref{tab:cec2013_so_funcs}), ranging from low-dimensional classical landscapes (1D--3D; e.g., Five-Uneven-Peak Trap, Equal/Uneven Maxima, Himmelblau, Six-Hump Camel Back, Shubert, and Vincent) to composite functions with increasing dimensionality up to 20D (F11--F20). 
These functions cover diverse configurations in terms of the number of global optima, the density of local optima (including ``Many''-peak cases), and peak-height profiles, thereby stress-testing an algorithm’s ability to locate and maintain multiple basins of attraction rather than producing pseudo-multimodality.

\begin{table}[H]
\centering
\footnotesize
\caption{CEC2013 single-objective multimodal benchmark functions (F1--F20).}
\label{tab:cec2013_so_funcs}
\setlength{\tabcolsep}{6pt}
\renewcommand{\arraystretch}{1.12}
\resizebox{\linewidth}{!}{%
\begin{tabular}{@{}l l r l r@{}}
\toprule
ID & Name & Global Optima & Local Optima & Peak Height \\
\midrule
F1  & Five-Uneven-Peak Trap (1D)     &   2  &  3    &  200.000 \\
F2  & Equal Maxima (1D)              &   5  &  0    &    1.000 \\
F3  & Uneven Decreasing Maxima (1D)  &   1  &  4    &    1.000 \\
F4  & Himmelblau (2D)                &   4  &  0    &  200.000 \\
F5  & Six-Hump Camel Back (2D)       &   2  &  4    &    1.032 \\
F6  & Shubert (2D)                   &  18  &  Many &  186.731 \\
F7  & Vincent (2D)                   &  36  &  0    &    1.000 \\
F8  & Shubert (3D)                   &  81  &  Many & 2709.094 \\
F9  & Vincent (3D)                   & 216  &  0    &    1.000 \\
F10 & Modified Rastrigin (2D)        &  12  &  0    &   -2.000 \\
F11 & Composite Function 1 (2D)      &   6  &  Many &    0.000 \\
F12 & Composite Function 2 (2D)      &   8  &  Many &    0.000 \\
F13 & Composite Function 3 (2D)      &   6  &  Many &    0.000 \\
F14 & Composite Function 3 (3D)      &   6  &  Many &    0.000 \\
F15 & Composite Function 4 (3D)      &   8  &  Many &    0.000 \\
F16 & Composite Function 3 (5D)      &   6  &  Many &    0.000 \\
F17 & Composite Function 4 (5D)      &   8  &  Many &    0.000 \\
F18 & Composite Function 3 (10D)     &   6  &  Many &    0.000 \\
F19 & Composite Function 4 (10D)     &   8  &  Many &    0.000 \\
F20 & Composite Function 4 (20D)     &   8  &  Many &    0.000 \\
\bottomrule
\end{tabular}}
\end{table}

\subsection{Decision-space basin coverage.}
\begin{figure*}[t]
  \centering
  \begin{subfigure}[b]{0.49\linewidth}
    \centering
    \includegraphics[width=\linewidth]{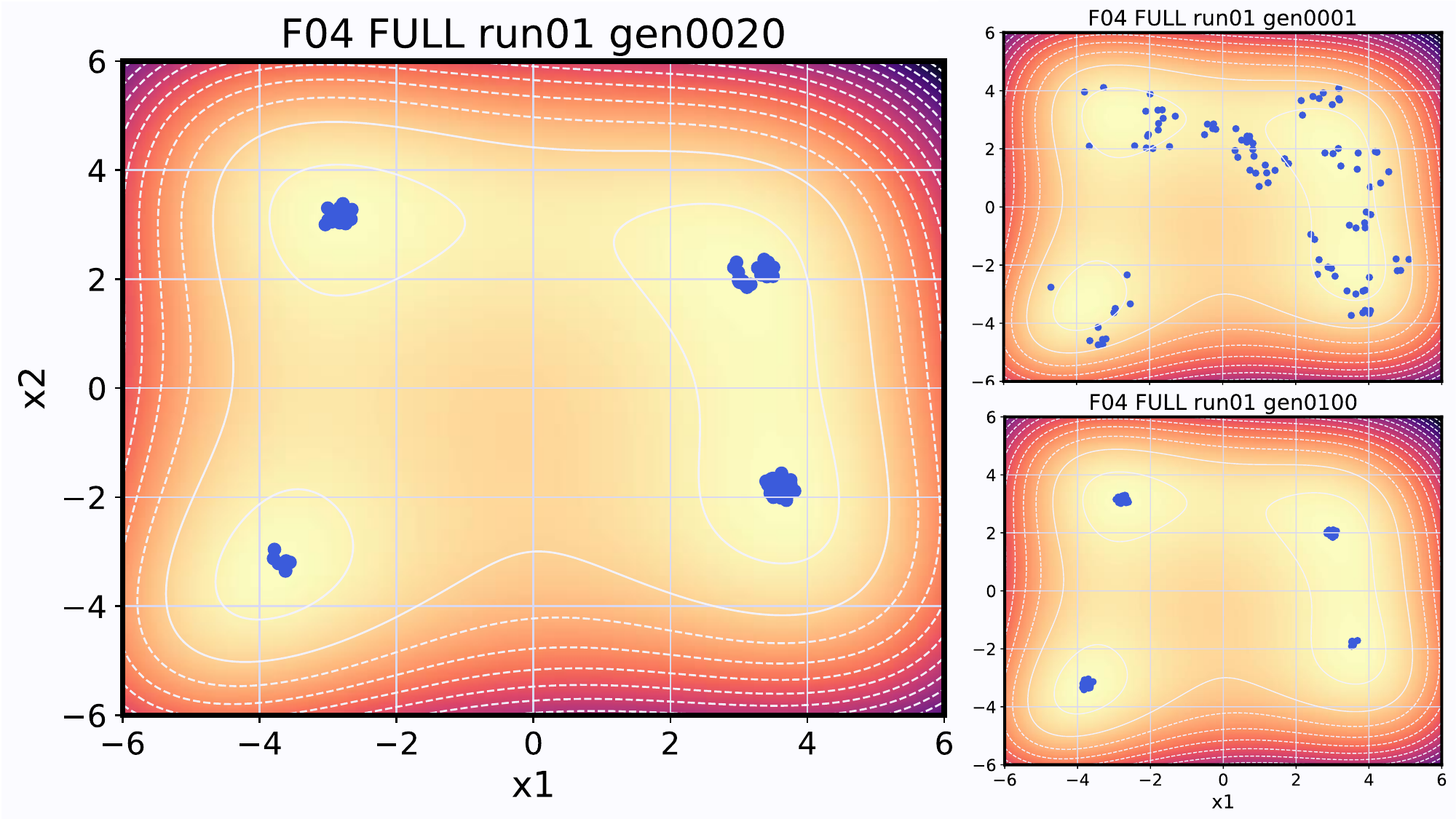}
    \caption{F04: the population splits into multiple stable basins and preserves distinct modes during refinement.}
    \label{fig:supp_peak_f04}
  \end{subfigure}\hfill
  \begin{subfigure}[b]{0.49\linewidth}
    \centering
    \includegraphics[width=\linewidth]{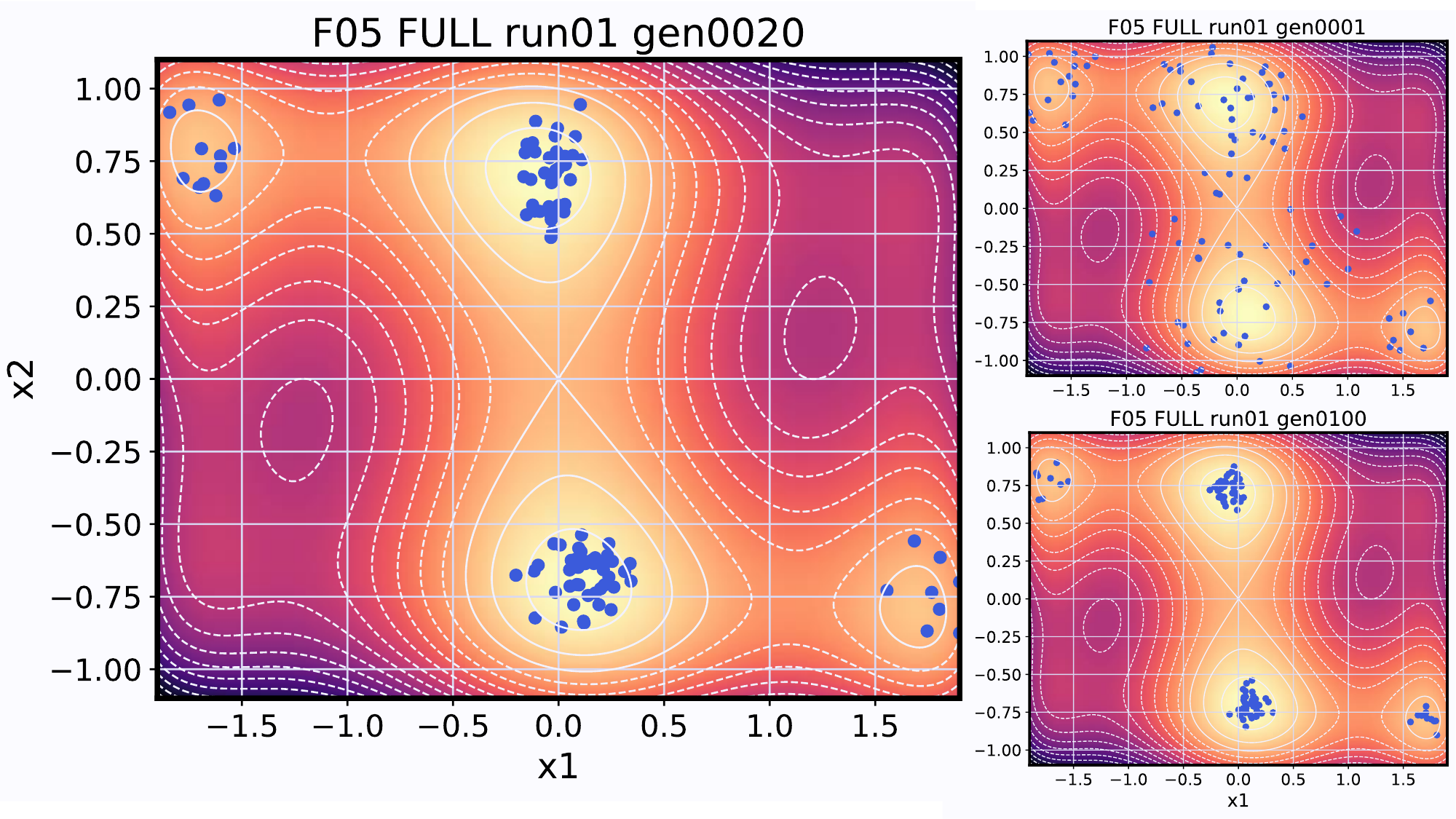}
    \caption{F05: rapid mode formation and basin-wise concentration after early exploration.}
    \label{fig:supp_peak_f05}
  \end{subfigure}

  \vspace{1mm}

  \begin{subfigure}[b]{0.49\linewidth}
    \centering
    \includegraphics[width=\linewidth]{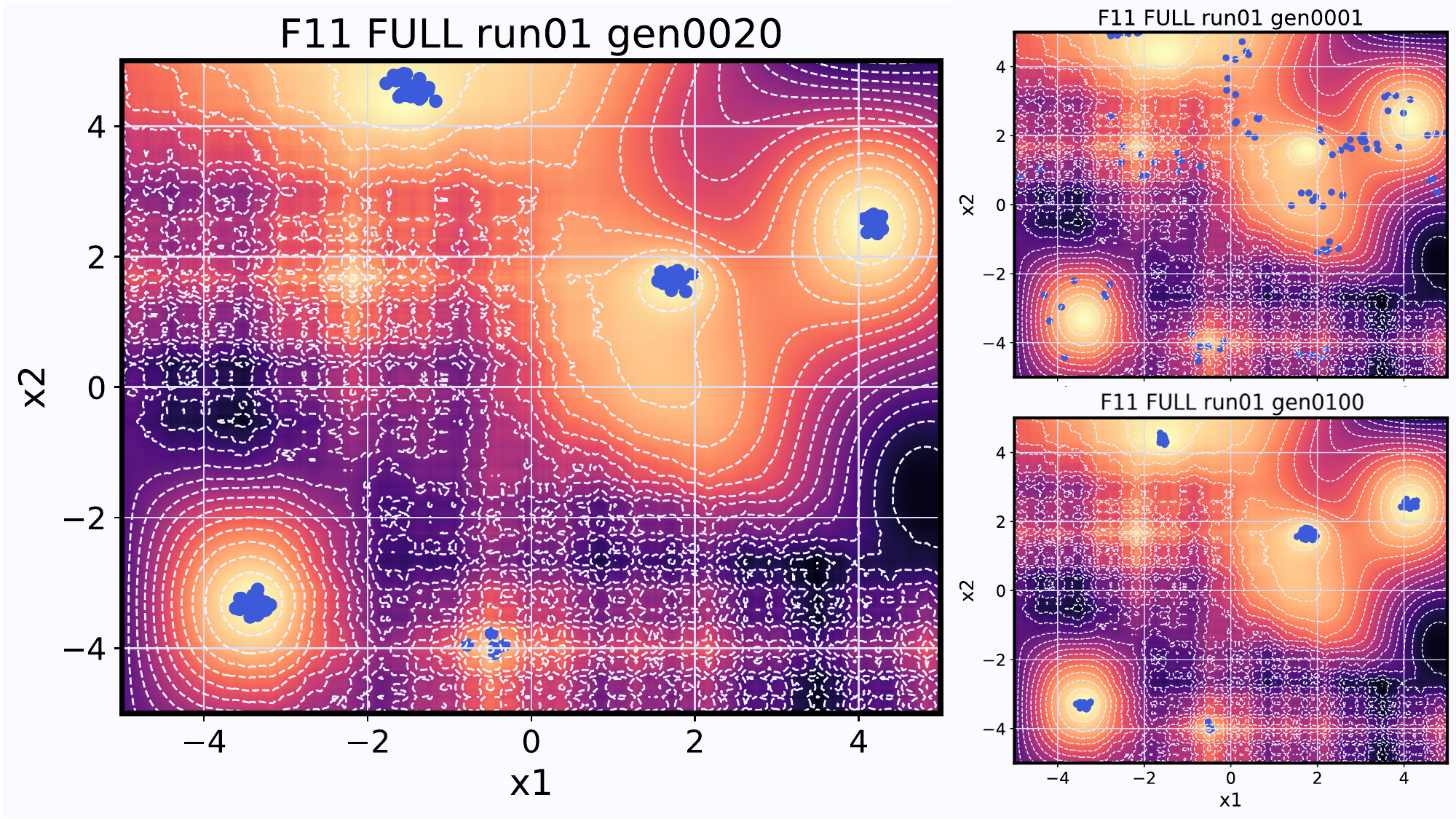}
    \caption{F11: on a rugged composite landscape, CLDE-S maintains several basin-wise clusters and avoids collapse.}
    \label{fig:supp_peak_f11}
  \end{subfigure}\hfill
  \begin{subfigure}[b]{0.49\linewidth}
    \centering
    \includegraphics[width=\linewidth]{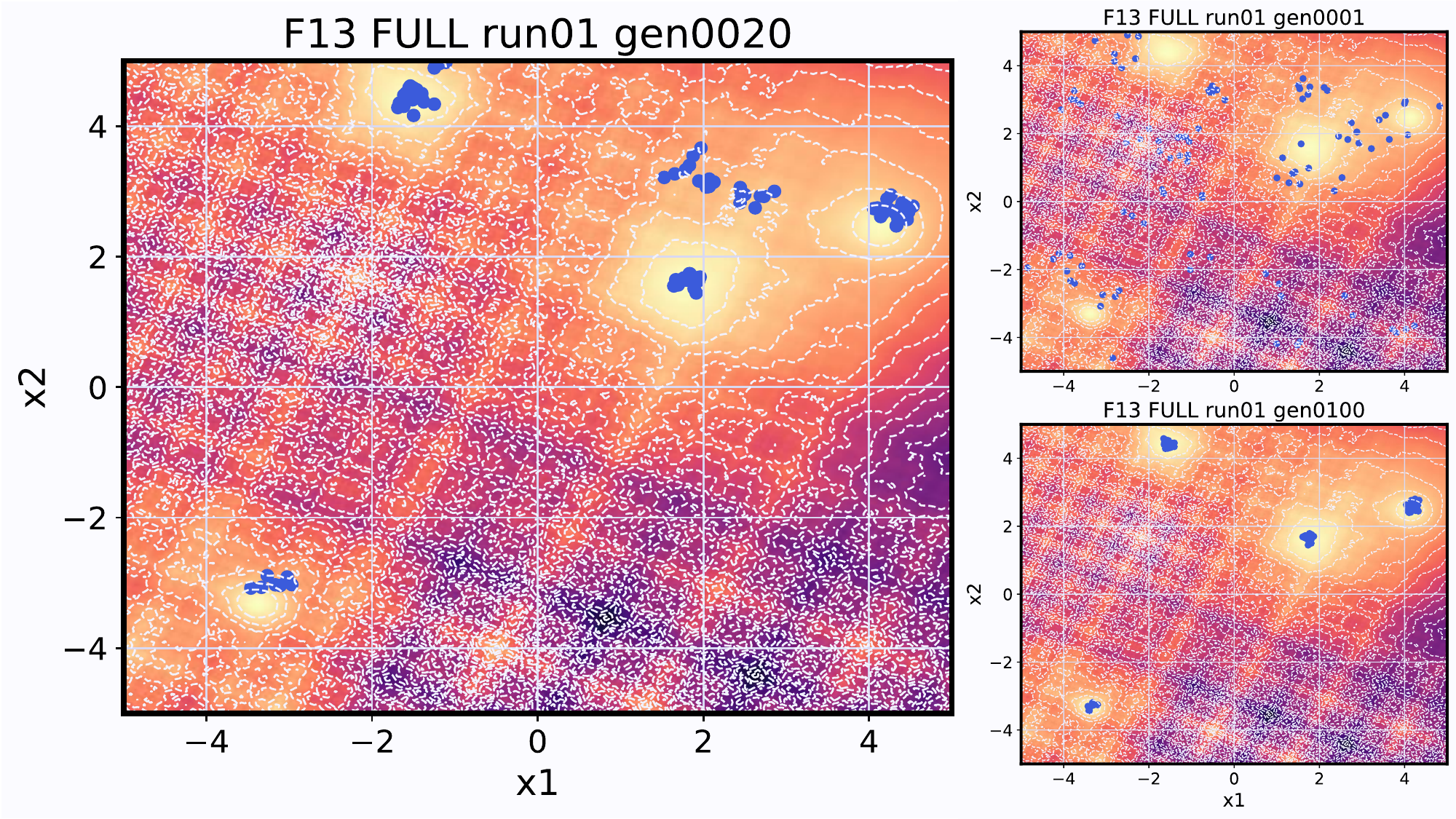}
    \caption{F13: multiple attraction regions are covered and progressively refined without collapsing to one basin.}
    \label{fig:supp_peak_f13}
  \end{subfigure}

  \caption{\textbf{(Figure \thefigure)} Population snapshots of CLDE-S on four representative 2-D CEC2013 functions.
  The contour background visualizes the objective landscape, and scatter points show the population in the decision space.
  Each panel contains a main snapshot at Gen 20, with insets for Gen 1 (early exploration) and Gen 100 (late refinement), illustrating the transition from basin discovery to basin-wise exploitation.}
  \label{fig:supp_peakplot}
\end{figure*}

To complement the PR comparisons and ablation results in the main paper, we provide qualitative decision-space evidence on how CLDE-S maintains multiple attraction basins over time.
Figure~\ref{fig:supp_peakplot} visualizes population snapshots on four representative 2-D CEC2013 functions (F04, F05, F11, and F13).
Across functions, CLDE-S quickly forms multiple basin-wise subpopulations in early generations and then refines them without collapsing to a single dominant mode, consistent with the intended exploration--decoding--allocation loop.
On rugged composite landscapes (e.g., F11 and F13), CLDE-S still preserves several coherent basin-wise clusters across generations, suggesting robust structural coverage under the same evaluation budget.
Together with the main PR and ablation results, these snapshots help explain why CLDE-S achieves stable peak maintenance rather than pseudo-diversity.

\section{Multi-Objective Multimodal Optimization}
\label{sec:supp_mo}

\noindent\textbf{Multi-objective multimodal benchmarks.}
For the multi-objective setting, we adopt representative DTLZ families and MMMOP problems.
The selected DTLZ instances (with $M=3$ objectives and decision dimensions 30/50) include both separable and non-separable cases and expose the optimizer to varying Pareto-front geometries (e.g., linear/concave, many-to-one, and multimodal variants such as DTLZ1/DTLZ3).
In addition, MMMOP benchmarks are designed to contain multiple equivalent and/or disconnected Pareto-optimal regions in the decision space (e.g., multiple or overlapping Pareto sets), which is particularly suitable for evaluating whether an algorithm can preserve multiple decision-space solution families beyond objective-space diversity.
All multi-objective benchmark problems and the corresponding performance evaluation in our experiments are implemented using the PlatEMO platform~\cite{PlatEMO}.

\begin{table}[ht]
\centering
\footnotesize
\caption{Multi-objective benchmark problems (DTLZ and MMMOP).}
\label{tab:mo_benchmarks}
\setlength{\tabcolsep}{3pt}
\renewcommand{\arraystretch}{1.05}
\begin{tabular}{@{}l c c c c l@{}}
\toprule
\textbf{Problem} & \textbf{M} & \textbf{Dim.} & \textbf{Separability} & \textbf{Modality} & \textbf{Geometry / Property} \\
\midrule
DTLZ1  & 3 & 30, 50 & Sep.     & M & Linear \\
DTLZ2  & 3 & 30, 50 & Non-sep. & U & Concave \\
DTLZ3  & 3 & 30, 50 & Sep.     & M & Concave \\
DTLZ4  & 3 & 30, 50 & Sep.     & U & Concave, many-to-one \\
DTLZ5  & 3 & 30, 50 & Sep.     & U & Concave \\
DTLZ6  & 3 & 30, 50 & Non-sep. & U & Concave \\
\midrule
MMMOP1 & 2 & 3 & Non-sep. & M & Multiple Pareto sets \\
MMMOP2 & 2 & 3 & Non-sep. & M & Multiple Pareto sets \\
MMMOP3 & 2 & 3 & Sep.     & M & Disconnected Pareto sets \\
MMMOP4 & 2 & 3 & Non-sep. & M & Overlapping Pareto sets \\
MMMOP5 & 2 & 3 & Sep.     & M & Multiple equivalent Pareto sets \\
MMMOP6 & 2 & 4 & Non-sep. & M & Multiple Pareto sets \\
\bottomrule
\end{tabular}
\end{table}

\paragraph{Decision-space statistical tests (IGD$_x$).}
To assess whether the observed decision-space advantages are statistically reliable, we perform Wilcoxon signed-rank tests on IGD$_x$ over MMMOP1--MMMOP6.
Table~\ref{tab:Wilcoxon_IGDX} reports the $p$-values between CLDE-M and competing algorithms.
Overall, the results provide additional statistical evidence that CLDE-M improves decision-space coverage on multiple MMMOP instances, consistent with the IGD$_x$ comparisons in the main paper.

\begin{table}[t]
\centering
\caption{\textbf{(Table \thetable)} Wilcoxon signed-rank test ($p$-values) between CLDE-M and competing algorithms on MMMOP1--MMMOP6 using IGD$_x$.}
\label{tab:Wilcoxon_IGDX}
\scriptsize
\setlength{\tabcolsep}{4pt}
\renewcommand{\arraystretch}{1.05}
\begin{tabular}{lccccc}
\toprule
\textbf{Problem} & \textbf{CMMO} & \textbf{CoMMEA} & \textbf{HREA} & \textbf{MMEAWI} & \textbf{MMEAPSL} \\
\midrule
MMMOP1 & 6.83e-2 & 5.74e-2 & 6.72e-2 & 4.56e-2 & 8.52e-2 \\
MMMOP2 & 6.04e-2 & 2.19e-2 & 1.30e-2 & 4.24e-2 & 3.24e-2 \\
MMMOP3 & 1.68e-2 & 4.40e-2 & 2.89e-2 & 1.98e-2 & 1.78e-2 \\
MMMOP4 & 2.91e-2 & 4.58e-2 & 2.22e-2 & 3.52e-2 & 1.70e-2 \\
MMMOP5 & 8.79e-3 & 5.72e-3 & 1.98e-2 & 4.10e-2 & 8.85e-3 \\
MMMOP6 & 2.80e-2 & 8.30e-3 & 2.37e-2 & 3.15e-2 & 2.88e-2 \\
\bottomrule
\end{tabular}
\end{table}

\paragraph{Ablation study on CLDE-M components (DTLZ1--DTLZ6).}
We further quantify the contribution of the main components in CLDE-M using an ablation study on DTLZ1--DTLZ6 with IGD.
Table~\ref{tab:ablation_dtlz} shows that the full CLDE-M consistently achieves the best IGD across all six problems.
Removing saliency-guided scoring, chaotic exploration, or basin-wise allocation degrades performance, indicating that these modules are complementary for sustaining exploration, maintaining multiple basins, and focusing evaluations on high-potential regions.

\begin{table}[t]
    \centering
    \caption{\textbf{(Table \thetable)} Ablation study of IGD on DTLZ1--DTLZ6 (lower is better).}
    \label{tab:ablation_dtlz}
    \scriptsize
    \setlength{\tabcolsep}{4pt}
    \renewcommand{\arraystretch}{1.05}
    \begin{tabular}{lccccc}
        \toprule
        \textbf{Function} & \textbf{NSGA-II} & \textbf{w/o Saliency} & \textbf{w/o Chaos} & \textbf{w/o BasinAlloc} & \textbf{CLDE-M} \\
        \midrule
        DTLZ1 & 22.50 & 21.80 & 19.40 & 18.10 & \textbf{17.55} \\
        DTLZ2 & 18.20 & 17.90 & 17.10 & 16.50 & \textbf{16.15} \\
        DTLZ3 & 23.15 & 22.05 & 20.25 & 18.80 & \textbf{18.20} \\
        DTLZ4 & 17.85 & 17.60 & 16.90 & 16.20 & \textbf{15.95} \\
        DTLZ5 & 15.40 & 15.10 & 14.80 & 14.25 & \textbf{13.90} \\
        DTLZ6 & 14.95 & 14.70 & 14.45 & 13.95 & \textbf{13.60} \\
        \bottomrule
    \end{tabular}
\end{table}

\section{Theoretical Notes}
\label{supp:theory}

\subsection{Time Complexity per Generation}
\label{supp:theory:complexity}

We analyze the computational cost of one generation of CLDE on a decoding canvas $S$ with $n=|S|$ points in $d$-dimensional decision space, using a $k$-nearest-neighbor (kNN) graph.

\paragraph{Step 1: Constructing the kNN graph.}
Let $G=(S,E)$ denote the kNN graph where each node connects to its $k$ nearest neighbors.
If distances are computed exhaustively, building kNN requires $O(n^2 d)$ distance evaluations, followed by selecting $k$ neighbors per node (which can be bounded by $O(n^2)$ using partial selection).
Thus, the overall worst-case time is
\[
T_{\text{kNN}} = O(n^2 d).
\]
In practice, approximate kNN or spatial indexing can reduce this cost, but the above worst-case bound holds independent of implementation.

\paragraph{Step 2: Basin decoding via union-find (disjoint set union).}
Given node heights $h(x)$ (fitness-based for SO or quality-based for MO), the decoding routine sorts nodes by height and processes them in descending order, maintaining connected components via union-find.
Sorting costs $O(n\log n)$, and union-find operations cost $O((n+|E|)\alpha(n))$, where $\alpha(\cdot)$ is the inverse Ackermann function.
Since $|E|=O(nk)$ for a kNN graph, the decoding time is
\[
T_{\text{decode}} = O(n\log n + nk\,\alpha(n)).
\]

\paragraph{Step 3: Basin scoring and budget allocation.}
Let $K$ be the number of decoded basins.
Computing basin-level statistics (e.g., best value, improvement, size) and assigning budgets can be implemented in $O(n)$ time by one pass over labels, plus $O(K)$ for normalization and quota calculation:
\[
T_{\text{score+alloc}} = O(n + K).
\]

\paragraph{Overall complexity.}
Therefore, the per-generation time complexity admits the bound
\[
T_{\text{gen}} = O(n^2 d) \;+\; O(n\log n + nk\,\alpha(n)) \;+\; O(n+K),
\]
where the kNN construction dominates in the worst case.
This bound is conservative and remains valid regardless of the specific benchmark suite or objective type.

\subsection{Correctness of Basin Decoding as Connected-Component Tracking}
\label{supp:theory:cc}

We provide a graph-theoretic interpretation of the basin decoding routine: it is equivalent to tracking connected components of a thresholded kNN graph induced by the height function.

\paragraph{Setup.}
Let $S$ be the decoding canvas and $G=(S,E)$ the kNN graph.
Let $h:S\to\mathbb{R}$ denote a scalar height (SO: fitness; MO: a scalar quality derived from rank/crowding or other monotone preference).
For any threshold $\lambda\in\mathbb{R}$, define the \emph{superlevel node set}
\[
S_\lambda = \{x\in S \mid h(x)\ge \lambda\},
\]
and the induced subgraph
\[
G_\lambda = G[S_\lambda].
\]
As $\lambda$ decreases, more nodes enter $S_\lambda$ and $G_\lambda$ grows.

\paragraph{Proposition 1 (Monotone growth).}
For $\lambda_1 \ge \lambda_2$, we have $S_{\lambda_1}\subseteq S_{\lambda_2}$ and $G_{\lambda_1}$ is a subgraph of $G_{\lambda_2}$.
Consequently, the number of connected components of $G_\lambda$ can only \emph{stay the same or decrease} as $\lambda$ decreases.

\paragraph{Proof.}
If $\lambda_1\ge \lambda_2$, then $h(x)\ge \lambda_1$ implies $h(x)\ge \lambda_2$, hence $S_{\lambda_1}\subseteq S_{\lambda_2}$.
The induced edges satisfy $E(G_{\lambda_1})\subseteq E(G_{\lambda_2})$.
Adding nodes/edges cannot create new connected components, so the component count is non-increasing. \qed

\paragraph{Proposition 2 (Equivalence to union-find decoding without merge gating).}
Consider the procedure that processes nodes in descending order of $h(x)$, and when a node $x$ is processed, it is activated and unioned with all already-activated neighbors in $G$.
Then, after processing all nodes with $h(x)\ge \lambda$, the union-find partition equals the set of connected components of $G_\lambda$.

\paragraph{Proof (sketch).}
By construction, the set of activated nodes after processing down to threshold $\lambda$ is exactly $S_\lambda$.
Union operations add precisely the edges in $G$ whose endpoints are both activated, i.e., edges of $G_\lambda$.
Union-find maintains the transitive closure of these edges, which is exactly the definition of graph connectivity in $G_\lambda$.
Thus the partition coincides with connected components of $G_\lambda$. \qed

\paragraph{Merge gating (as used in our decoding).}
Our decoding additionally applies a merge-gating rule (controlled by $\tau$) to selectively prevent certain component merges even if a connecting node exists.
This can be viewed as replacing $G$ by a \emph{pruned graph} $\tilde{G}=(S,\tilde{E})$ where an edge (or a merge event) is retained only if it passes the gating criterion.
All statements above remain valid if $G$ is replaced by $\tilde{G}$, i.e., the decoding still corresponds to connected-component tracking on a well-defined thresholded graph $\tilde{G}_\lambda$.
Therefore, the decoding procedure has a clear graph-theoretic meaning and does not rely on ambiguous geometric assumptions.

\section{Chaotic Sensitivity: $\mu=4$}
\label{supp:chaos_mu}

\begin{figure}[htbp]
  \centering
  \includegraphics[width=0.98\linewidth]{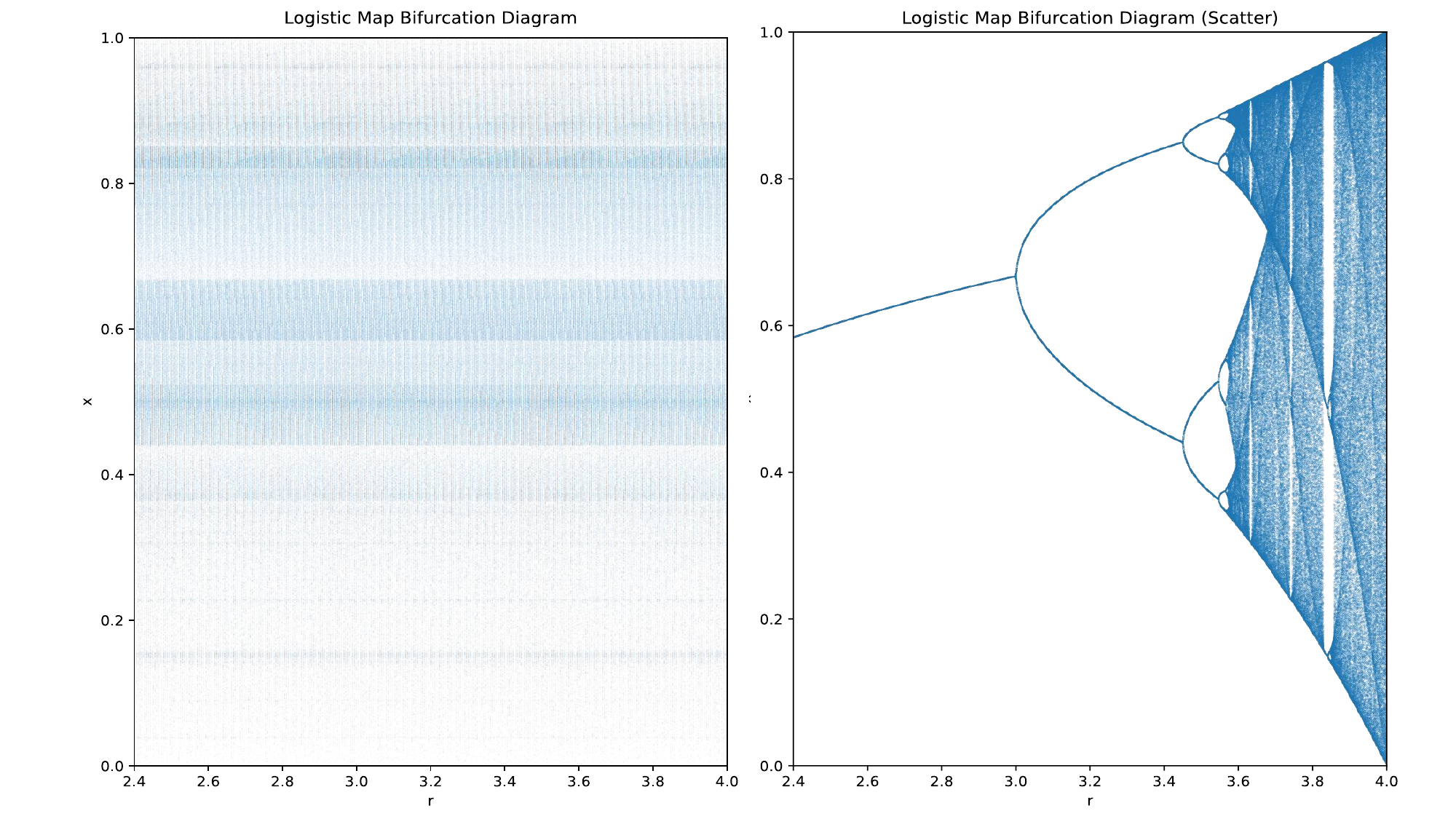}
  \caption{Bifurcation diagram of the logistic map $x_{t+1}=\mu x_t(1-x_t)$. As $\mu$ increases, the system transitions from stability ($\mu<3$) to period-doubling ($3\le\mu<3.57$) and then to fully developed chaos ($\mu\gtrsim 3.57$). The fully chaotic regime provides quasi-random and decorrelated perturbations, motivating the use of $\mu=4$ as a strong exploration setting in the CE module.}
  \label{fig:bifurcation-pair}
\end{figure}

To justify the default chaotic intensity adopted in the Chaotic Evolution (CE) module (Sec.~3.1), we conducted a supplementary sensitivity study on the logistic-map parameter $\mu$ in Eq.~3. 
The objective is to verify whether increasing the nonlinearity level can strengthen global exploration while maintaining stable convergence under a fixed evaluation budget.

\begin{figure*}[htbp]
  \centering
  \includegraphics[width=0.9\linewidth]{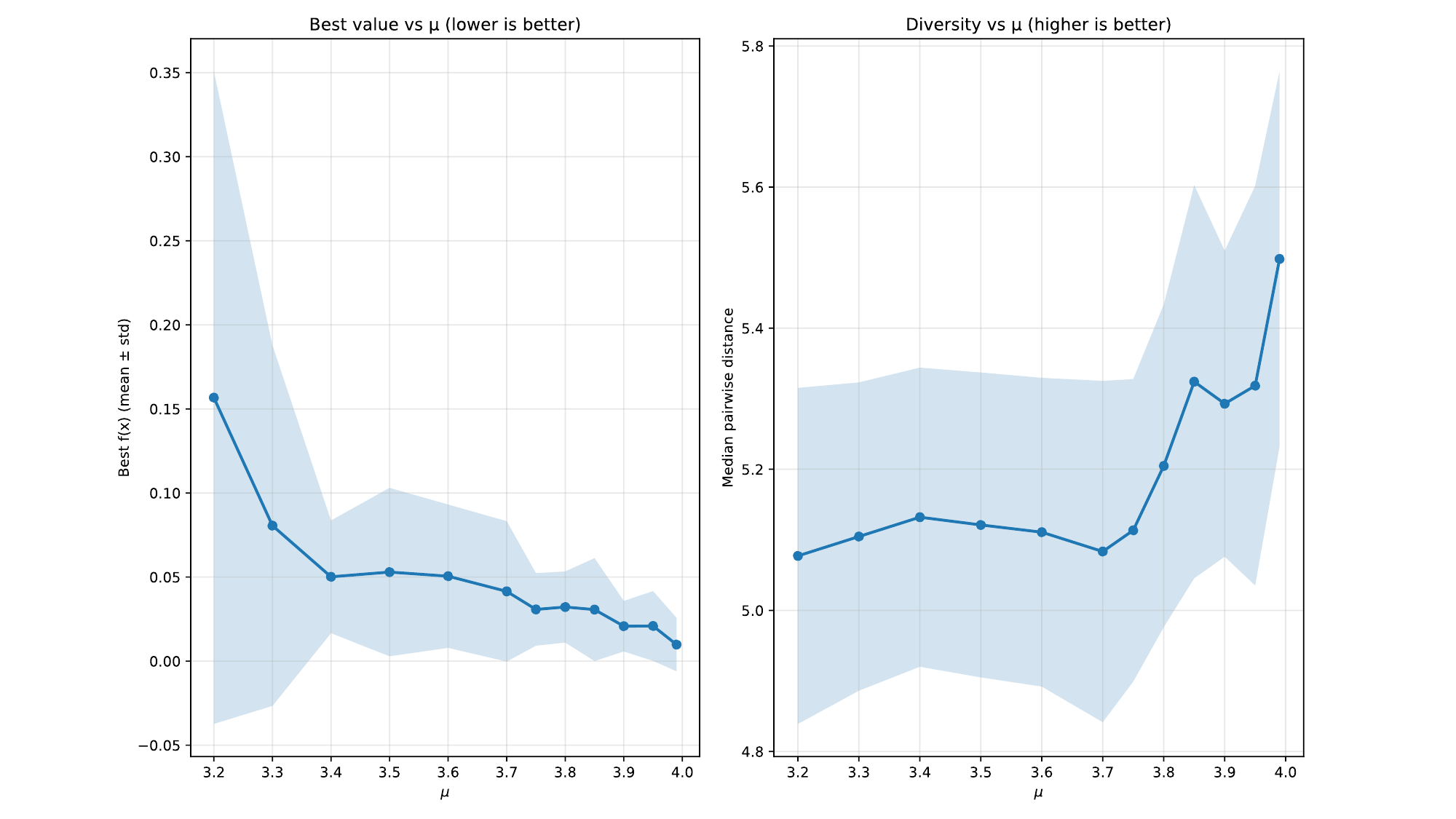}
  \caption{Effect of the chaos parameter $\mu$ on optimization performance and diversity. The left panel reports the final best objective value (lower is better) and the right panel reports the median pairwise distance (higher is better), averaged over repeated runs under the same evaluation budget. The period-doubling range yields unstable performance and reduced diversity, whereas the fully chaotic regime improves both. Among all tested settings, $\mu=4$ achieves the lowest final objective value while maintaining the highest diversity, supporting our default choice.}
  \label{fig:mu-metrics}
\end{figure*}

\subsection{Regime interpretation}
Figure.~\ref{fig:bifurcation-pair} shows the bifurcation diagram of the logistic map $x_{t+1}=\mu x_t(1-x_t)$. 
As $\mu$ increases, the dynamics transition from stability ($\mu<3$) to period-doubling ($3\le \mu < 3.57$) and then to the fully chaotic regime ($\mu \gtrsim 3.57$).
The period-doubling range produces structured and correlated sequences, which can result in repetitive perturbation patterns and limited diversity. 
In contrast, the fully chaotic regime yields quasi-random fluctuations that are more suitable for broad exploration and for sustaining population diversity, motivating our focus on $\mu\in(3.5,4)$.

\subsection{Empirical impact on performance and diversity}
We evaluate the effect of $\mu$ using two complementary indicators in Figure.~\ref{fig:mu-metrics}: 
(i) the final best objective value (lower is better) and 
(ii) the median pairwise distance in the population (higher indicates stronger diversity), both computed over repeated runs under the same protocol.
The period-doubling region exhibits unstable optimization outcomes together with reduced diversity, consistent with the correlated nature of the perturbations.
After the onset of chaos ($\mu>3.57$), diversity increases and performance becomes markedly more stable.
Notably, among the tested settings, $\mu=4$ achieves the best overall trade-off by simultaneously attaining the lowest final objective value and the highest diversity in our experiments (Figure.~\ref{fig:mu-metrics}).

\subsection{Why we set $\mu=4$}
Based on the above observations, we use $\mu=4$ as the default.
First, $\mu=4$ corresponds to the standard upper bound of the logistic map in the fully chaotic regime and produces the strongest decorrelated perturbations, which promotes basin escape and prevents premature convergence. 
Second, in our experiments, $\mu=4$ consistently yields the best final solution quality while maintaining the largest population diversity (Figure.~\ref{fig:mu-metrics}), providing direct empirical support for this choice.
As a result, $\mu=4$ offers a robust and implementation-simple default for the CE module.

\end{document}